\PassOptionsToPackage{table}{xcolor}
\documentclass[acmsmall]{acmart}

\AtBeginDocument{%
  \providecommand\BibTeX{{%
    \normalfont B\kern-0.5em{\scshape i\kern-0.25em b}\kern-0.8em\TeX}}}

\setcopyright{acmcopyright}
\copyrightyear{2021}
\acmYear{2021}
\acmDOI{10.1145/3472752}

\graphicspath{ {./figures/} }

\usepackage{subcaption}
\usepackage{longtable}
\usepackage{lscape}
\usepackage[table]{xcolor}
\usepackage{todonotes}

\begin{document}

\title{A Survey on Concept Drift in Process Mining}

\author{Denise Maria Vecino Sato}
\email{denise.sato@ppgia.pucpr.br}
\email{denise.sato@ifpr.edu.br}
\orcid{0000-0003-1117-7082}
\author{Sheila Cristiana de Freitas}
\email{freitas.sheila@ppgia.pucpr.br}
\email{sheila.freitas@ifpr.edu.br}
\affiliation{%
  \institution{Graduate Program in Informatics (PPGIa), Pontifícia Universidade Católica do Paraná}
  \streetaddress{R. Imac. Conceição, 1155}
  \city{Curitiba}
  \state{Paraná}
  \country{Brazil}
  \postcode{80215-901}
}
\affiliation{%
  \institution{Instituto Federal do Paraná}
  \streetaddress{R. João Negrão, 1285}
  \city{Curitiba}
  \state{Paraná}
  \country{Brazil}
  \postcode{80230-150}
}

\author{Jean Paul Barddal}
\email{jean.barddal@ppgia.pucpr.br}
\author{Edson Emilio Scalabrin}
\email{scalabrin@ppgia.pucpr.br}
\affiliation{%
  \institution{Graduate Program in Informatics (PPGIa), Pontifícia Universidade Católica do Paraná}
  \streetaddress{R. Imac. Conceição, 1155}
  \city{Curitiba}
  \state{Paraná}
  \country{Brazil}
  \postcode{80215-901}
}

\renewcommand{\shortauthors}{Sato, et al.}

\begin{abstract}
Concept drift in process mining (PM) is a challenge as classical methods assume processes are in a steady-state, i.e., events share the same process version. We conducted a systematic literature review on the intersection of these areas, and thus, we review concept drift in process mining and bring forward a taxonomy of existing techniques for drift detection and online process mining for evolving environments. Existing works depict that (i) PM still primarily focuses on offline analysis, and (ii) the assessment of concept drift techniques in processes is cumbersome due to the lack of common evaluation protocol, datasets, and metrics.
\end{abstract}



\maketitle

\section{Introduction: Background and motivation}
\label{sec:Introduction}

Process Mining (PM) is a growing research area that provides techniques to understand and improve processes in different application domains \cite{Garcia2019}. PM's goal is to extract non-trivial process-related information from event data (observed behavior) recorded by the information systems available. The PM area is drawing more attention because of the increasing availability of data events (more data is being generated and recorded) and the need to develop and improve business processes in fast-changing environments. In \cite{VanderAalst2016a}, van der Aalst defines three types of PM: discovery, conformance, and enhancement. Discovery aims at producing a process model that generalizes the observed behavior from the event data without any \emph{a priori} information. The discovered process model may describe only the control-flow of events or other aspects, e.g., organizational or time. In conformance, the goal is to compare the event data with a process model (discovered or designed) to reveal discrepancies. Enhancement focuses on improving or extending the current process model based on information obtained from event data. The key elements for any PM technique are the event data and the process model.

\begin{figure}[hbt!]
  \centering
  \includegraphics[width=0.8\linewidth]{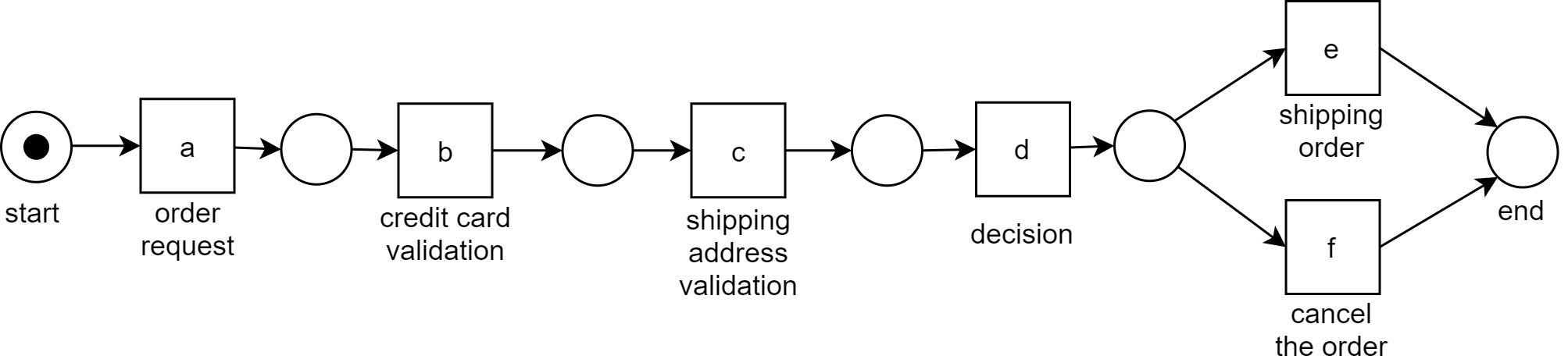}
  \caption{Petri net model for the e-commerce process.}
  \Description{A process model from an e-commerce application describing the flow of activities after the order request.
  The process contains the following activities: order request (a), credit card validation (b), shipping address validation (c), decision (d), shipping order (e), and cancel the order (f). Activities a to d are sequentially executed, then there is a choice between e and f.}
  \label{fig:ecommerceprocess1}
\end{figure}

The event data are usually recorded by information systems and can be accessed in event logs or event streams. An event log contains a historical record of occurred events for a specific process, providing information for the PM techniques in an offline manner. Event streams allow the PM techniques to operate online, accessing the events as they occurred. When comparing event streams to event logs, the main differences are that the event stream is potentially infinite, and the cases inside it can be incomplete \cite{VanZelst2018a}. Regardless of the format, the event data contain information about the execution of a single business process, e.g., buying items online. The process model in Figure \ref{fig:ecommerceprocess1} shows the sequence of activities executed in an online store using the Petri Net notation \cite{VanderAalst2016a}.

\begin{figure}[hbt!]
  \centering
  \includegraphics[width=0.8\linewidth]{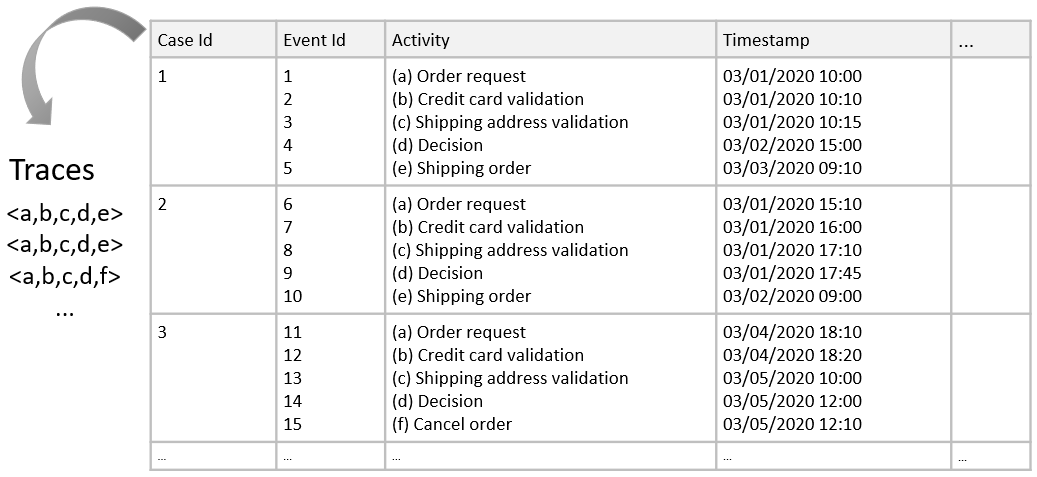}
  \caption{Example of event data.}
  \Description{Event data example related to the e-commerce process. Three cases are described by their sequence of activities.
  Cases 1 and 2 generate the trace <a,b,c,d,e> and case 3 generate the trace <a,b,c,d,f>.}
  \label{fig:eventdata}
\end{figure}

Each event refers to a single process instance (case), is linked to some activity of the process (e.g., order request), and must be ordered within a case. Figure \ref{fig:eventdata} shows an example of event data recorded for the e-commerce process (Figure \ref{fig:ecommerceprocess1}). Each case has an identifier and represents one execution of the process, recording what is called a trace in the event data, i.e., the sequence of events for the specific case. In the described cases, the trace \(<a,b,c,d,e>\) is recorded for cases 1 and 2; but case 3 generates a different trace \(<a,b,c,d,f>\). The event data in Figure \ref{fig:eventdata} contains the minimum information needed for PM techniques (case identifier, activity, and timestamp), but other case or event attributes can be recorded and included in the process analysis. Because several cases can record the same trace, a simple representation of the event log contains the traces and its frequency, denoted as a superscript: \(L=[<a,b,c,d,e>^2,<a,b,c,d,f>^1,...]\).

Most existing PM methods are designed to work with event logs and consider the processes in a steady-state, i.e., all traces are related to the same version of the process. However, real-life processes are likely to change over time in response to market dynamics, new regulations, or even improving or repairing the current process. This scenario imposes a new challenge: dealing with these changes in the process analysis. A change inside a process can be planned or unexpected, and the situation where the process is changing while being analyzed is named concept drift \cite{Bose2014}. 

Concept drift is a relevant problem in different domains, e.g., business process analysis. Companies are always trying to adapt and evolve their business process to handle different situations, e.g., changes in regulations and seasonal demands, so PM techniques should consider the concept drift challenge to allow process analysis in evolving business. The PM manifesto \cite{VanderAalst2011} also highlights dealing with concept drift in PM as one of the core challenges for the area. However, there is no structured review of the topic describing the available published methods. This survey covers different aspects of dealing with concept drift in PM, providing an overview of the current techniques and the challenges as a contribution to the area.   

The paper is organized as follows. In Section \ref{sec:Concept drift in process mining}, we introduce the concept drift in the PM context, presenting its main challenges. Sections \ref{sec:Concept drift detection} and \ref{sec:Online PM dealing with evolving environments} overview current methods that can be applied to evolving processes. Section \ref{sec:Achievements and challenges} discusses the main achievements of the existing approaches and highlights open challenges. Finally, Section \ref{sec:Conclusion} concludes the survey.   

\section{Concept drift in process mining}
\label{sec:Concept drift in process mining}

Concept drift is a well-known problem in data mining, and it refers to an online supervised learning scenario when the relation between the input data and the target variable changes over time \cite{Lu2019,Ditzler2015}. In this situation, the learning algorithms need to adapt themselves on the fly to react to concept drifts appropriately \cite{Gama2014}. In PM, we also have a relationship between the input data, i.e., the event data, and the target variable, i.e., the process model. Nevertheless, data mining techniques, including techniques to deal with concept drift, handle structures as the target variable, as categorical or continuous values in the form of a vector as input. PM techniques deal with process models, which contain more complex structures, like concurrency, choices, and loops.
Therefore, it is not possible to use the same methods developed for data mining in PM, and thus, it motivates the development of new strategies and techniques to handle concept drift in PM. To understand the proposed approaches for dealing with concept drift in PM, we conducted a Systematic Literature Review (SLR) to sustain the discoveries of this survey.

\subsection{Research method}

The SLR's goal is to identify the current approaches for dealing with concept drift in PM. We only included peer-reviewed documents and defined the following research questions (RQ):

\begin{itemize}
    \item RQ1: Which approaches authors proposed to deal with concept drift detection in processes?
    \item RQ2: Which approaches provide tools for concept drift detection in processes?
    \item RQ3: How did the authors validate the proposed approaches?
\end{itemize}

The first author executed the search protocol by applying the search strings in the databases of four digital libraries, as described in Table \ref{tab:slr_search_strings}. We have also included the workshops from the International Conference on PM Series (ICPM) because of their relevance to the PM area.

\begin{table}[hbt!]
  \caption{Searches in Digital Libraries and proceedings from ICPM Workshops.}
  \label{tab:slr_search_strings}

  \begingroup
  \renewcommand{\arraystretch}{1.3} 
  \resizebox{0.75\textwidth}{!}{
  \begin{tabular}{p{0.2\textwidth}p{0.6\textwidth}p{0.1\textwidth}}
    \toprule
    Origin&Search String&Results\\
    \midrule
    ACM Digital Library & [[All: ``concept drift''] OR [All: ``process drift'']] AND [All: ``process mining''] & 20\\
    IEEE Xplore & ((``All Metadata'':``concept drift'' OR ``process drift'') AND ``All Metadata'':``process mining'') AND [All: ``process mining''] & 13\\
    Scopus & ( TITLE-ABS-KEY ( ``concept drift''  OR  ``process drift'' )  AND  TITLE-ABS-KEY ( ``process mining'' ) )  AND  ( LIMIT-TO ( LANGUAGE ,  ``English'' ) ) AND [All: ``process mining''] & 56\\
    Springer Link & (``concept drift'' OR ``process drift'') AND ``process mining'' - LIMITED TO ENGLISH & 140\\
    ICPM Workshops & ``concept drift''  & 2\\
  \midrule
  Total & & 231 \\
  \bottomrule
  \end{tabular}
  }
  \endgroup
  
\end{table}

After obtaining the first set of papers from the digital libraries, the next step was to read and classify them based on the inclusion (IC) and exclusion criteria (EC). The ICs define the relevant studies, while the ECs guarantee the quality of the selected ones based on the pre-defined RQs \cite{Kitchenham2007}:

\begin{itemize}
  \item IC1: Peer-reviewed documents written in English describing approaches for dealing with concept drift in process models using only event data as input - 88 papers.
  \item IC2: Peer-reviewed documents written in English comparing approaches or tools for dealing with concept drift in process models - 2 papers.
  \item EC1: Duplicated documents - 35 papers.
  \item EC2: Papers published before 2015 with less than five citations - 5 papers.
  \item EC3: Approaches dealing with concept drift in a business process that do not consider changes in a complete perspective of the model, e.g., control-flow or time. For instance, a predictive model for a specific time difference in the process, e.g., delivery time as the target adapted when a concept drift occurred - 6 papers.
  \item EC4: Approaches for dealing with concept drift in processes but reporting experiments without complete detailing, e.g., without describing the dataset configuration - 2 papers.
\end{itemize}

Finally, we applied backward snowballing \cite{Wohlin2014} while following the ICs and ECs, including three papers summing up a total of 45 papers. After reading the selected papers, we organized an overview of the concept drift in PM, thus providing its main concepts. Sections \ref{sec:Concept drift detection} and \ref{sec:Online PM dealing with evolving environments} describe the identified approaches for dealing with concept drifts.

\subsection{Concept drift definition}

Concept drift research in PM focused on two directions: (i) detecting drifts and providing information for its analysis, and (ii) offering online PM techniques for dealing with evolving environments. 
We have also identified two papers \cite{Omori2019,Ceravolo2020} comparing approaches for concept drift detection in process models, which are included in the first branch.

\begin{figure}[hbt!]
  \centering
  \includegraphics[width=0.65\linewidth]{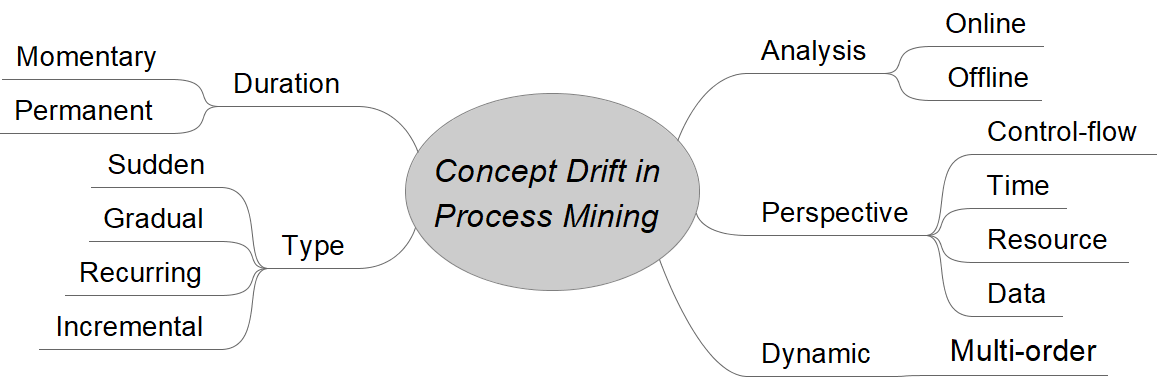}
  \caption{Different aspects of defining and characterizing concept drift in process mining.}
  \Description{The concept drift in process mining is defined and characterized by the
  following aspects: duration (momentary or permanent), type (sudden, gradual, recurring, incremental),
  analysis (online, offline), perspective (control-flow, time, resource, data), dynamic (multi-order).}
  \label{fig:conceptdriftPM}
\end{figure}

The concept drift in PM, also named business process drift in \cite{Maaradji2015}, is the situation when the process changes while being analyzed \cite{VanderAalst2016a}. Therefore, we cannot assume that the event log or stream contains traces of a unique version of the process. Process models may include different perspectives of the process: control-flow, data, timing, or others, and the changes can also occur in any of these perspectives. Several notations exist to model business processes, and we will refer to them as process models. The concept drift in process models includes different aspects reported in the related works, summarized in Figure \ref{fig:conceptdriftPM}. The following topics illustrate and explain the aspects depicted above.

\subsection{Perspectives of change}

A process model typically contains information about different perspectives; therefore, the changes can occur in one or more of them. Authors in \cite{Bose2011,Bose2014} highlighted three essential perspectives in business processes: control-flow, data, and resource (also named as organizational perspective by \cite{VanderAalst2016a}). Another vital view in PM is the time perspective, which provides performance indicators as lead, waiting, service, and synchronization times. Despite this perspective being widely explored for process analysis, we found fewer papers exploring it for concept drifts \cite{Brockhoff2020,Richter2017,Richter2019,BarbonJunior2018,Mora2020,Tavares2019}.

\subsubsection{Control-flow}

\begin{figure}[hbt!]
  \centering
  \begin{subfigure}{0.7\linewidth}
  \includegraphics[width=\linewidth]{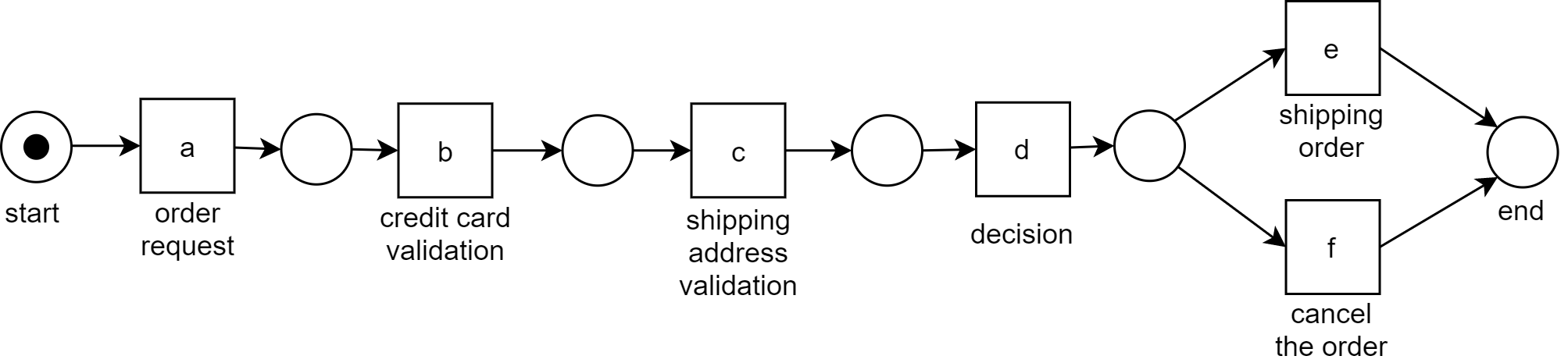} 
  \caption{Process model before the drift.}
  \label{fig:ECommerceProcessA}
  \end{subfigure}
  \\
  \begin{subfigure}{0.7\linewidth}
  \includegraphics[width=\linewidth]{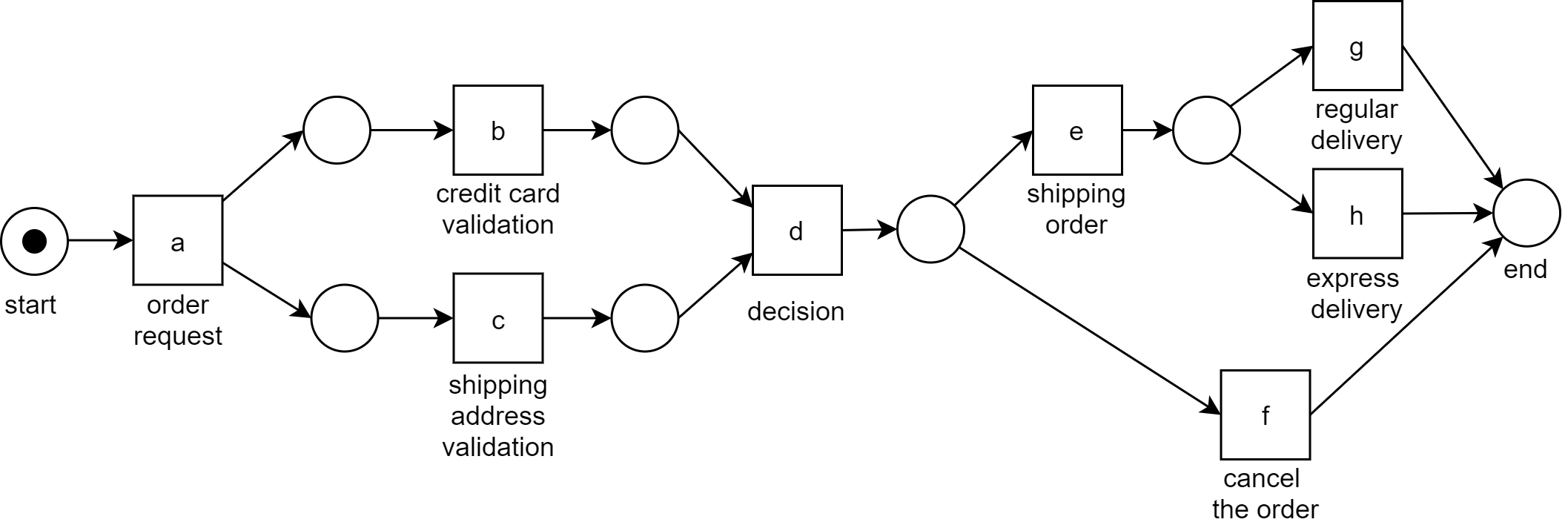}
  \caption{Process model after the drift.}
  \label{fig:ECommerceProcessB}
  \end{subfigure}

  \caption{Concept drift in the control-flow perspective.}
  \Description{Two process models from an e-commerce application describing the flow of activities before and after a concept drift. In the first process, the activities a and d are sequentially executed, then there is a choice between activities e and f. In the second version of the process model, activities b and c can now be executed in parallel. It is included a choice between two new activities, g and h, after the activity e.}
  \label{fig:ECommerceProcessStructuralChange}
  \centering
\end{figure}

Changes in this perspective represent behavioral or structural changes in the process model. Process models can be imperative, e.g., Petri nets or BPMN, or declarative, e.g., Declare models \cite{VanDerAalst2009a}. Authors in \cite{Weber2008} suggest a set of 13 structural change patterns for imperative business process models, organized into adding/deleting fragments, moving/replacing fragments, adding/removing levels (sub-processes), adapting control dependencies, and changing transition conditions. In a declarative process model, a structural change can be adding or deleting a constraint. Figure \ref{fig:ECommerceProcessStructuralChange} shows two structural changes in the e-commerce process: two activities for delivery are included within a choice construction (regular or express), and the activity \emph{shipping address validation} can now be executed in parallel with \emph{credit card validation}.

The process behavior can also change without any structural modification on the model. In the e-commerce process example, customers can be classified as \emph{silver} or \emph{gold}, depending on their purchase history. \emph{Gold} customers may start having express deliveries without extra charges. The process structure remains intact, but the routing of cases changes as all orders from \emph{gold} customers are routed to the \emph{express delivery} activity. 

\subsubsection{Time}

This perspective concerns the timing and frequency of events in the process. Events recorded with timestamps make it possible to identify bottlenecks or measure service levels \cite{VanderAalst2016a}. A change in this perspective deals with significant changes in process performance over time \cite{BarbonJunior2018,Tavares2019,Mora2020,Brockhoff2020}. An example can be an activity that is manually executed, e.g., credit card validation, and after a specific date is changed to be automatically performed by the system. The automatic activity can reduce its time of execution, changing the time perspective of the process. 

\subsubsection{Resource}

A change in the resource perspective regards the influence of the resources on the execution of activities, including changes in the organization structure, roles, and resource availability. An activity may not be active until a specific resource, e.g., a machine on a production line, is available. A situation like this can change the execution paths of the process. In another process, some activities can be redirected to be executed by a different department. In the e-commerce process example, we can change the department responsible for the \emph{decision} activity, for instance, changing the resource perspective. 

\subsubsection{Data}

This perspective concerns the data produced or consumed by the process during the execution of its activities. A change in this perspective represents a change in the data related to the case or the event. In the e-commerce process example, it may no longer be required to attach the credit card image to execute the \emph{credit card validation} activity. 
The proposals depicted in \cite{Hompes2017,Pauwels2019,Stertz2019} explicitly deal with this perspective.

\subsection{Duration, type, and dynamic of the change}

Changes can be classified as momentary or permanent, depending on the period the change is active. Momentary changes are short-lived and affect only a few cases, whether permanent changes are persistent and stay for a while \cite{Schonenberg2008}. In PM, a momentary change can be understood as an outlier (or blip) that represents an unusual behavior. Typically, PM techniques filter out the outliers. Thus, the approaches to deal with concept drift usually focus only on permanent drifts. Authors in \cite{Bose2011,Bose2014} identified four distinct types of drifts: sudden, gradual, recurring, and incremental. 

\subsubsection{Type of Drift}

\begin{figure}[hbt!]
  \centering
  \includegraphics[width=\linewidth]{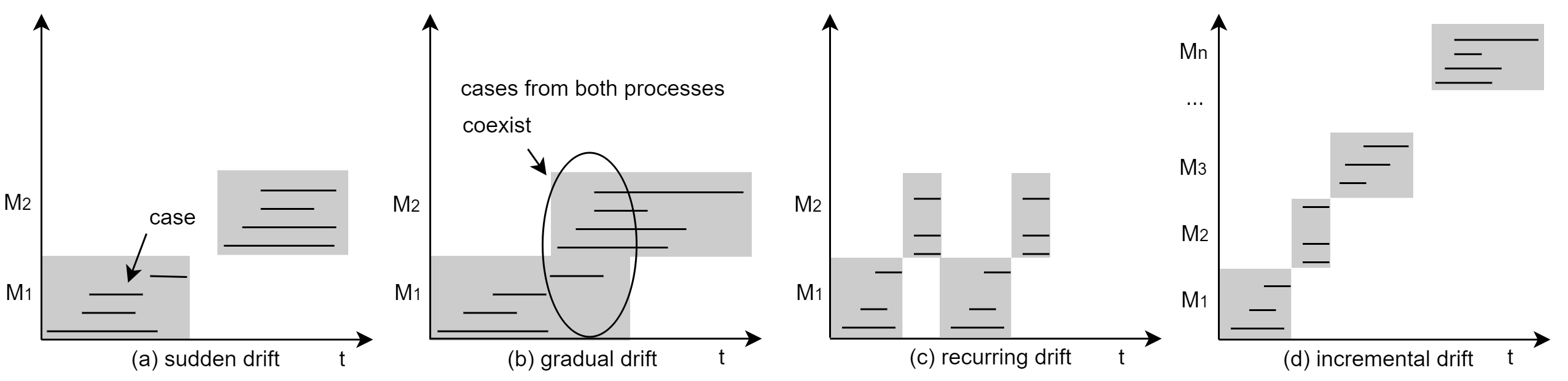}
  \caption{Four different types of drift. Adapted from \cite{Bose2014}.}
  \Description{The four types of drifts (sudden, gradual, recurring, and incremental) are represented over time. In the sudden type, the first version of the process is abruptly changed. In the gradual type, there is a period where both versions of processes co-exist. In the recurring type, the initial
  version of the process comes back. In the incremental type, several small changes occurred over time.}
  \label{fig:drifttypes}
\end{figure}

Figure \ref{fig:drifttypes} shows: (a) a sudden drift, (b) a gradual drift, (c) a recurring drift, and (d) an incremental drift. 
The x-axis represents the time component, while the y-axis indicates distinct process models.
A line inside the shaded rectangles represents a case. 

\paragraph{Sudden drift}

When a sudden drift occurs, a new version of the process model starts to handle all the ongoing cases. It represents a complete substitution of the current process \(M_{1}\) with a new one \(M_{2}\), as shown in Figure \ref{fig:drifttypes} (a). This type of drift can occur in emergencies or even when new regulations must be followed. In a sudden drift, incomplete cases should be redirected to the new process, especially in real situations; this is also named intra-trace drift by \cite{Ostovar2016}. In this situation, a case can contain events related to different versions of the process. Several approaches handle sudden drifts, but the synthetic logs used for validation do not always contain intra-trace drifts.   

\paragraph{Gradual drift}

\begin{figure}[hbt!]
  \centering
  \includegraphics[width=\linewidth]{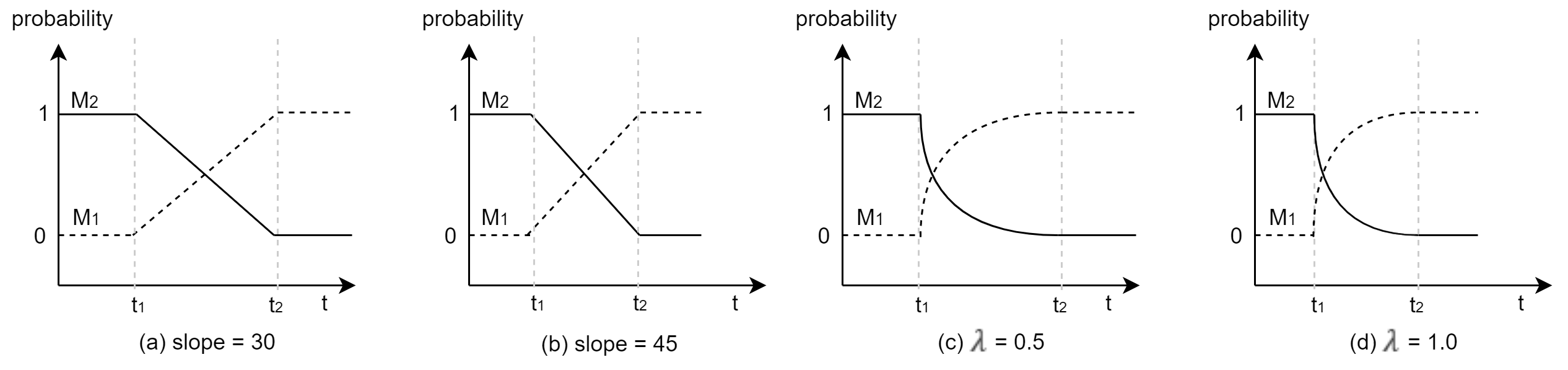}
  \caption{(a) and (b) shows variants of gradual linear drift, with slope defining the rate of change. (c) and (d) shows variants of exponential gradual drift, characterized by the function \(e^{-\lambda t}\) for \(M_{1}\). Adapted from \cite{Martjushev2015}.}
  \Description{Two variants of gradual linear drift, with slopes 30 and 45 defining the rate of change. Two variants of exponential gradual drift, characterized by the function \(e^{-\lambda t}\), with \(\lambda 0.5\) and \(1.0\) }
  \label{fig:gradualdriftvariants}
\end{figure}

A gradual drift occurs when the current process \(M_{1}\) is replaced by a new one \(M_{2}\), but instances of both processes coexist for a while. For instance, if only the new customers of the e-commerce process should provide a photo of the identification document. As indicated in Figure \ref{fig:drifttypes} (b), after some time, only instances of \(M_{2}\) will exist. Figure \ref{fig:gradualdriftvariants} illustrates different gradual drifts showing the probability of a process model emanating instances in the y-axis and time in the x-axis. We can artificially model gradual drifts in different ways by using distinct functions to describe how things grow or decay as time passes. Figure \ref{fig:gradualdriftvariants} (a-b) shows a gradual drift characterized by a linear change between \(M_{1}\) and \(M_{2}\), i.e., the cases from both processes continuously decrease and increase. The slope defines the degree of decrease/increase (Figure \ref{fig:gradualdriftvariants} a–b), and after \(t_{2}\), all cases emanate only from \(M_{2}\). Another common approach is to follow an exponential rate of increase/decrease of cases from two processes, such as depicted in Figure \ref{fig:gradualdriftvariants} (c–d). 
In this example, between $t_1$ and $t_2$, which define the beginning and the end of the drift region, $M_1$ emanates cases with \(P[M_1] = e^{-\lambda t}\) probability and $M_2$ follows $P[M_2] = (1 - P[M_1])$ \cite{Martjushev2015}.

\paragraph{Recurring drift}

Recurring drifts represent situations where a process \(M_{1}\) is replaced by the process \(M_{2}\), but after some time, \(M_{1}\) reappears, as in Figure \ref{fig:drifttypes} (c). If a model, or a set of models, reappears after a while, we have a recurrent drift. 
It is also noteworthy to mention that models' replacements in recurring drifts can be either sudden or gradual.
Furthermore, a process that changes based on seasonal influences is an example of a recurring drift. 

\paragraph{Incremental drift}

Incremental drifts represent the situation where a model \(M_{1}\) is replaced to the model \(M_{n}\) by incremental minor changes, as illustrated in Figure \ref{fig:drifttypes} (d). As in the recurring drift, each change inside an incremental drift can represent a sudden or gradual drift. \\

Summing up, the sudden and gradual drifts can be considered basic patterns of change. Recurring and incremental can combine gradual, sudden, or even both types of drifts. In Figure \ref{fig:drifttypes}, they are shown as discrete sudden changes. 

\subsubsection{Dynamic of change}

In \cite{Martjushev2015} authors described multi-order dynamic, i.e., a situation where process changes happen at different time granularity levels.
For instance, let us assume a problem where a company changes its process after a four week span.
Nonetheless, during such a period, the company has two variants of the process both before and after the drift.
Approaches for dealing with concept drift in PM must consider multi-order dynamics, which indicates a different time scale for the changes. In the example, we can identify micro-level changes every week and macro-level changes after four weeks. This situation is complicated when the approaches use fixed-time windows. Adaptive time windows can be more suitable for this scenario. However, in both cases, the window size hyperparameter affects drift identification. 

\subsection{Type of analysis}

PM techniques can be executed either online or offline. In an offline analysis, data are collected in event logs, representing historical information from a time period and afterward analyzed. Most PM techniques are designed for offline analysis, and concept drift is also explored in this scenario. Possible applications for offline concept drift analysis are: (i) splitting the event log into smaller parts for a better understanding of the process, thus avoiding more complex and imprecise models; (ii) indicating unknown and probably unexpected changes in the processes, improving the process analysis; and (iii) using the results of the analysis for redesigning the processes. In an offline setting, the time for drift detection is not a vital issue, and thus, the approaches can consider data after the drift on the analysis, and the event logs can be filtered to maintain only complete traces. 

An online analysis (also known as operational support \cite{VanderAalst2016a}) provides a way to influence and to react to the ongoing cases by accessing the events as they occurred, usually as event streams, which are an infinite sequence of events generated over time \cite{VanZelst2018a}. An online setting usually requires some assumptions inherited from the data mining domain: data should have a fixed and small number of attributes; algorithms should be able to process unlimited data without exceeding memory restrictions; algorithms should consider a finite amount of available memory in a reasonable time; there is a small upper bound on time allowed to process an event, e.g., usually algorithms work with one pass of the data; and stream 
``concepts'' may be stationary or evolving \cite{Buttazzo2011}. We need to address the concept drift problem in online mode when the presence of changes or the occurrence of drifts needs to be uncovered in real-time or in near real-time. The online analysis is appropriate if the organization is interested in reacting to a change when it is happening, using real-time alarms. For concept drift detection in PM, this means identifying the process changes as soon as possible but with confidence that the change is significant. Therefore, we can highlight two main constraints for online drift detection: accuracy and time. The online methods should handle both constraints to find a good trade-off between them. 

The approaches to online detect concept drift usually define a period where events are collected to determine a reference model. After defining the reference model, the drift mechanism starts to process new events. Concept drift detection approaches should not store all events from the stream, so it is essential to define a forgetting mechanism. The methods can forget events by adopting a windowing strategy  (considering the recent events on the analysis) or applying additional methods, e.g., the aging factor. Some authors proposed concept drift online detection using a stream of traces \cite{Liu2018,Maaradji2015}. However, we only considered online strategies when event streams are the input, as a stream of traces requires waiting for traces to be complete before its inclusion into the stream. 

\subsection{Dealing with concept drifts in processes}

We identified two branches in the 45 identified papers: (i) concept drift detection; and (ii) online PM dealing with evolving environments (Table \ref{tab:papers_classification}). Sections \ref{sec:Concept drift detection} and \ref{sec:Online PM dealing with evolving environments} detail the approaches.

\begin{table}[hbt!]
  \caption{Classification of papers dealing with concept drift in PM.}
  \label{tab:papers_classification}
  \begingroup
  \renewcommand{\arraystretch}{1.3} 
  \resizebox{0.75\textwidth}{!}{
  \begin{tabular}{p{0.3\linewidth}p{0.45\linewidth}p{0.18\linewidth}}
    \toprule
    Approach & Papers & Number of papers\\
    \midrule
    Concept drift detection & \cite{Bose2011,Accorsi2012,Carmona2012,Luengo2012,Bose2014,Maaradji2015,ManojKumar2015,Martjushev2015,Ostovar2016,Hompes2017,Maaradji2017,Ostovar2017,Richter2017,Seeliger2017,Zheng2017,BarbonJunior2018,Liu2018,Stertz2018,Hassani2019,KurniatiA.P.McInerneyC.ZuckerK.HallG.HoggD.2019,Pauwels2019,Omori2019,Richter2019,Stertz2019,Tavares2019,Yeshchenko2019,Yeshchenko2019a,Brockhoff2020,Ceravolo2020,Impedovo2020,Kurniati2020,Lin2020,Mora2020,Ostovar2020,Richter2020,Yeshchenko2020,Zellner2020,Yeshchenko2021} 
    & 38\\
    Online PM dealing with evolving environments & \cite{Maggi2013,Burattin2014,Redlich2014,Burattin2015a,Burattin2015b,VanZelst2018a,Batyuk2020} & 7\\
  \bottomrule
  \end{tabular}}
  \endgroup
\end{table}

\section{Concept drift detection}
\label{sec:Concept drift detection}

The approaches for concept drift detection usually addressed one or more challenges described in \cite{Bose2011,Bose2014}. 
We differentiate drift detection from change point (CP) detection because some approaches only detect the drift without reporting the CP \cite{Carmona2012,Hassani2019,KurniatiA.P.McInerneyC.ZuckerK.HallG.HoggD.2019,Kurniati2020}. 
We also separate ``Change localization and characterization'', defined in \cite{Bose2011,Bose2014}, as such approaches usually focus on localization. 

\begin{enumerate}
    \item \textbf{Drift detection.} Detects that a process has changed without providing exact information about the time period or the trace/event the change occurred. 

    \item \textbf{Change point detection.} Detects that a process has changed and identifies the time period or the exact point (event or trace) where the drift occurred. These approaches usually report the case or event identifier as the CP, but we have also identified one approach \cite{Yeshchenko2019,Yeshchenko2019a,Yeshchenko2020,Yeshchenko2021}, named VDD, which reports the day as the CP.

    \item \textbf{Change localization.} Identifies the region(s) of change in the process model.  The method that deals with change localization should identify the exact point inside the model where the drift occurs, e.g., between activities \emph{A} and \emph{B}, without needing a process model as input. 

    \item \textbf{Change characterization.} Characterizes the nature of the change, defining the perspective of change and the type of drift, e.g., sudden or gradual. As several approaches deal with a specific perspective and type, e.g., sudden drifts in the control-flow perspective, this challenge is unusually reported.

    \item \textbf{Unravel process evolution or change process discovery.} Reveals the complete change process based on the identification, localization, and characterization of a change. Process analysts need tools to explore and relate all the discoveries, resulting in discovering the change process describing the drift dynamics. 
\end{enumerate}

\begin{figure}[hbt!]
  \centering
  \includegraphics[width=0.65\linewidth]{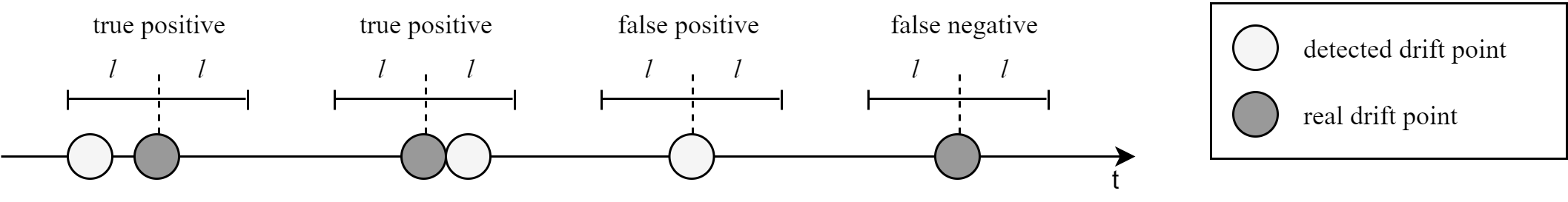}
  \caption{Definition of evaluation metrics. Adapted from \cite{Martjushev2015}.}
  \Description{The figure shows the method considers a true positive when there is a detected CP in the interval of size \(l+2\) around a real drift point; a false positive when the method detects a drift and in the interval of size \(l+2\) around the detected drift there is no real drift; and a false negative when there is no detected drift point in the interval of size \(l+2\) around a real drift point.}
  \label{fig:evaluationmetrics}
\end{figure}

Table \ref{tab:drift_detection_papers} provides an overview of the drift detection approaches in PM. If the link of the public dataset is not available anymore, we do not classify it as publicly available. When the authors validate the approaches using synthetic datasets, we also reported if an objective metric is calculated. We identified two metrics in the papers: \emph{F-score} and \emph{detection delay}. The \emph{F-score} is the most common metric and reports the geometric mean of the \emph{precision} and \emph{recall}, which rely on true positives (TP), false positives (FP), and false negatives (FN). Yet, the definition of a (\emph{TP}), (\emph{FP}), and (\emph{FN}) is unclear in 14 papers. In Table \ref{tab:drift_detection_papers} only the papers with the \emph{F-score} clearly defined are marked with \emph{CD}. For instance, if the method returns the trace index as the CP, a \emph{TP} can consider a range of indexes around the actual drift, and this should be specified. In \cite{Martjushev2015,Lin2020}, the authors clearly define the evaluation mechanism (Figure \ref{fig:evaluationmetrics}). The difference from other papers is that the authors apply a lag period \emph{l} surrounding a detected or an actual drift. In \cite{Zheng2017}, the same idea is applied with the name Error Tolerance (\emph{ET}). Techniques able to detect drifts with high precision and recall are preferred over others. The \emph{detection, average or mean delay} indicates the average number of traces/events processed by the method between the actual drift and the moment the drift is alerted \cite{Maaradji2015}. It points out how early the approach is able to detect an actual change \cite{Seeliger2017}. 
For the sake of homogeneity, hereafter, we use the \emph{detection delay} term.

The authors in \cite{Luengo2012} report the percentage of correctly classified traces based on the known process version. The approach is based on trace clustering, and the idea is to verify if the clustering strategy correctly classifies the traces. It is not exactly a generic metric, but it works on trace clustering approaches. In \cite{Tavares2019}, the authors compare the total number of detected drifts with the actual number of drifts. We do not consider this an accurate metric because it can evaluate detection accuracy with a bias; however, we reported it in Table \ref{tab:drift_detection_papers}. The authors in \cite{Impedovo2020} report an accuracy value (from 0 to 1) without defining the metric, so we did not include it in Table \ref{tab:drift_detection_papers}.

We only considered online approaches those that use event streams as input. The authors in \cite{Carmona2012} propose a method that can be applied online. However, the validation applied used a stream of traces. Each trace is converted into several Parikh vectors (complete traces, one by one). The Parikh vectors should be created based on an event stream for online use. The method proposed in \cite{Liu2018} is classified as offline because the event stream defined is, in fact, a stream of traces.

We grouped the papers by approach, and, in this case, we identified the characteristics addressed in some of the papers using citation. The two papers comparing tools for drift detection are not included and are described in Section \ref{sec:Comparing tools for drift detection}. The papers are organized by the year of publication and title (when there is more than one paper, the ordering considers the first one), in ascending order.

Table \ref{tab:drift_detection_papers} highlights that synthetic datasets containing artificially injected drifts are scarce. We only found three publicly available datasets that we further detail in Appendix A.

We organized the approaches to detect concept drift in processes based on the strategy for drift detection: statistical hypothesis testing, trace clustering, visual analysis, CP detection, change detection, trend detection, and other approaches. We classified the 36 papers in Table 4, and the most common approach is statistical hypothesis testing. 

The two papers \cite{Omori2019,Ceravolo2020} that compare drift detection approaches are excluded from this classification. 
It is important to highlight that some of the proposed methods combine other strategies before or after applying the drift detection, e.g., clustering.

The following sections describe each approach in detail. Different aspects from the distinct addressed challenge(s) addressed by each approach are also described. Section \ref{sec:Comparing tools for drift detection} describes the two papers that compare drift detection approaches.

\subsection{Statistical hypothesis testing}

Several papers use statistical-based approaches to detect concept drifts in processes. In this section, we group these approaches by the challenge addressed and the type of drift detected.

\subsubsection{Detecting sudden drifts}

The majority of the papers bring forward techniques that apply statistical hypothesis testing to identify significant changes in the event data, process model, or both over time, thus indicating a potential concept drift. These approaches assume that there should be a statistically significant difference in traces, process models, or both; before and after the CPs \cite{Bose2011,Bose2014}. The main idea is to apply a hypothesis test to confirm if there is statistical evidence indicating the two samples are equal or not. Most of the approaches using statistical hypothesis testing are tailored to deal with the first two challenges of process drift detection: drift detection and CP detection. 

Figure \ref{fig:statisticalhypothesistesting} shows an overview of the papers' steps to explain the approach to detect and pinpoint a process drift (challenges 1 and 2 from Section \ref{sec:Concept drift detection}) based on statistical hypothesis testing. 
The grey rectangles are the main steps for any standard statistical hypothesis testing that is based on event data. 
The options for each step were described based on the papers identified in this survey.

\begingroup

\definecolor{lightgray}{gray}{0.9}
 
\begin{landscape}
\rowcolors{1}{}{lightgray}
{\begin{longtable}{p{2.1cm}p{1.8cm}p{2.3cm}p{2.3cm}p{2.2cm}p{3cm}p{1.8cm}p{2.3cm}}
  \caption{Drift detection papers in PM grouped by approach. SD means synthetic datasets, RD means real-world datasets, PA means publicly available, NA means not available, and CD means clearly defined.} \label{tab:drift_detection_papers} \\

  \toprule \multicolumn{1}{l}{Paper(s)} & \multicolumn{1}{l}{Challenge(s)} & \multicolumn{1}{l}{Perspective(s)} & \multicolumn{1}{l}{Type(s)} & \multicolumn{1}{l}{Analysis} & \multicolumn{1}{l}{Software} & \multicolumn{1}{l}{Dataset} & \multicolumn{1}{l}{SD evaluation}\\ \midrule
  \endfirsthead

  \multicolumn{8}{l}%
  {{\tablename\ \thetable{} -- continued from previous page}} \\
  \toprule \multicolumn{1}{l}{Paper(s)} & \multicolumn{1}{l}{Challenge(s)} & \multicolumn{1}{l}{Perspective(s)} & \multicolumn{1}{l}{Type(s)} & \multicolumn{1}{l}{Analysis} & \multicolumn{1}{l}{Software} & \multicolumn{1}{l}{Dataset} & \multicolumn{1}{l}{SD evaluation}\\ \midrule
  \endhead

  \midrule \rowcolor{white!100} \multicolumn{8}{r}{{Continued on next page}} \\ \midrule
  \endfoot

  \bottomrule
  \endlastfoot
 
\cite{Bose2011,Bose2014,Martjushev2015} & 1,2,3 & Control-flow & Sudden, Gradual\cite{Bose2014,Martjushev2015} & Offline & ProM (ConceptDrift) & SD, RD (PA) & F-score CD\cite{Martjushev2015}\\

\cite{Luengo2012} & 3 & Control-flow & Gradual & Offline & Tool NA & SD & \% of correctly classified traces\\

\cite{Accorsi2012} & 1,2,3 & Control-flow & Sudden & Offline & Tool NA & SD & -\\

\cite{Carmona2012} & 1 & Control-flow & Sudden & Online & Tool NA & SD & -\\

\cite{ManojKumar2015} & 1,2 & Control-flow & Sudden & Offline & Tool NA & SD (PA) & -\\

\cite{Maaradji2015,Ostovar2016,Maaradji2017,Ostovar2017,Ostovar2020} & 1,2,3 & Control-flow & Sudden, Gradual\cite{Maaradji2017} & Offline\cite{Maaradji2015,Maaradji2017}, Online\cite{Ostovar2016,Ostovar2017,Ostovar2020} & Apromore (ProDrift) & SD (PA), RD (PA) & F-score, detection delay\\

\cite{Hompes2017} & 1,2 & Control-flow, Data & Sudden & Offline & ProM (TraceClustering) & SD, RD (PA) & -\\

\cite{Seeliger2017} & 1,2,3 & Control-flow & Sudden & Offline & Experimental ProM plug-in\footnotemark & SD (PA) & F-score, detection delay 
\footnotetext{Source code available at https://github.com/alexsee/processdriftdetector.}\\

\cite{Zheng2017} & 1,2 & Control-flow & Sudden & Offline & TPCDD\footnotemark & SD & F-score CD
\footnotetext{Source code available at https://github.com/THUBPM/process-drift-detection. Event logs are not made available.}\\

\cite{Richter2017,Richter2019} & 1,2,3 & Time & Sudden, Incremental, Recurring & Online & ProM (Tesseract) & SD, RD (PA) & Detection delay\\

\cite{BarbonJunior2018,Tavares2019,Mora2020} & 1,2 & Control-flow and time combined & Not defined, Sudden \cite{Tavares2019} & Online & CDESF\footnotemark & SD (PA), RD (PA) & N. of detected drifts \cite{Tavares2019}
\footnotetext{Source code available at https://github.com/gbrltv/CDESF \cite{BarbonJunior2018}. Updated version available in https://github.com/gbrltv/cdesf2 \cite{Tavares2019,Mora2020}.}\\

\cite{Liu2018} & 1,2 & Control-flow & Sudden, Recurring & Offline & Tool NA & RD (PA) & F-score\\

\cite{Stertz2018,Stertz2019} & 1,2 & Control-flow \cite{Stertz2018}, data\cite{Stertz2019} & Sudden, Gradual, Incremental, Recurring & Online & Tool NA & SD & -\\

\cite{KurniatiA.P.McInerneyC.ZuckerK.HallG.HoggD.2019,Kurniati2020} & 1 & Control-flow & Not defined & Offline & Manual method & RD & -\\

\cite{Pauwels2019} & 1,2 & Control-flow and data combined & Not defined & Offline & EDBN\footnotemark & RD (PA) & -
\footnotetext{Source code available at https://github.com/StephenPauwels/edbn.}\\

\cite{Hassani2019} & 1 & Control-flow & Sudden & Online & StrProMCDD NA & SD, RD (PA) & F-score, detection delay\\

\cite{Yeshchenko2019,Yeshchenko2019a,Yeshchenko2020,Yeshchenko2021} & 1,2,3,4 \cite{Yeshchenko2020,Yeshchenko2021} & Control-flow & Sudden, Gradual, Incremental, Recurring & Offline & VDD\footnotemark & SD (PA), RD (PA) & F-score \cite{Yeshchenko2019, Yeshchenko2021}\footnotetext{Source code available at https://github.com/yesanton/Process-Drift-Visualization-With-Declare. Web client available at https://yesanton.github.io/Process-Drift-Visualization-With-Declare/client/build/.}\\

\cite{Zellner2020} & 1,2 & Control-flow & Recurring & Offline & DOA\footnotemark & SD, RD (adapted) & F-score\footnotetext{Source code available at https://github.com/zellnerlu/DOA.}\\

\cite{Richter2020} & 1,2 & Control-flow & Not defined & Online & OTOSO\footnotemark & RD (PA) & -\footnotetext{Source code available at https://github.com/Skarvir/OTOSO.}\\

\cite{Brockhoff2020} & 1,2 & Control-flow and time combined & Sudden & Offline & ProM & SD & -\\

\cite{Lin2020} & 1,2 & Control-flow & Sudden & Offline & LCDD\footnotemark & SD (PA), RD (PA) & F-score CD\footnotetext{Source code available at https://github.com/lll-lin/THUBPM.}\\

\cite{Impedovo2020} & 1,2,3 & Control-flow & Sudden & Offline & Tool NA & SD, RD (PA) & -\\
\end{longtable}}
\end{landscape}

\begin{table}[hbt!]
  \caption{Classification of papers dealing with concept drift detection in PM based on the detection approach.}
  \label{tab:detection_approaches}
  \begingroup
  \renewcommand{\arraystretch}{1.3} 
  \resizebox{0.75\textwidth}{!}{
  {\begin{tabular}{lll}
    \toprule
    Approach for Drift Detection & Papers & Number of papers\\
    \midrule
    Statistical hypothesis testing & \cite{Bose2011,Bose2014,Maaradji2015,ManojKumar2015,Martjushev2015,Ostovar2016,Maaradji2017,Ostovar2017,Seeliger2017,Pauwels2019,Ostovar2020} & 11 \\
    
    Trace clustering & \cite{Accorsi2012,Luengo2012,Richter2017,BarbonJunior2018,Tavares2019,Mora2020,Zellner2020} & 7 \\
    
    CP detection & \cite{Yeshchenko2019,Yeshchenko2019a,Yeshchenko2020,Yeshchenko2021} & 4 \\
    
    Visual analysis & \cite{Hompes2017,KurniatiA.P.McInerneyC.ZuckerK.HallG.HoggD.2019,Brockhoff2020,Kurniati2020} & 4 \\

    Change detection & \cite{Carmona2012,Hassani2019,Impedovo2020} & 3 \\
    
    Trend detection & \cite{Richter2017,Richter2019} & 2 \\
    
    Other approaches & \cite{Zheng2017,Liu2018,Stertz2018,Stertz2019,Lin2020} & 5 \\
    
  \bottomrule
  \end{tabular}}}
  \endgroup
\end{table}

Figure \ref{fig:statisticalhypothesistesting} overviews the possible combinations that a method can choose to identify process drifts. 
The following topics describe each of the main steps explaining how the authors apply the proposed approach. Finally, Table \ref{tab:statistical} details the choice for each step per paper.

\begin{figure}[hbt!]
  \centering
  \includegraphics[width=0.7\linewidth]{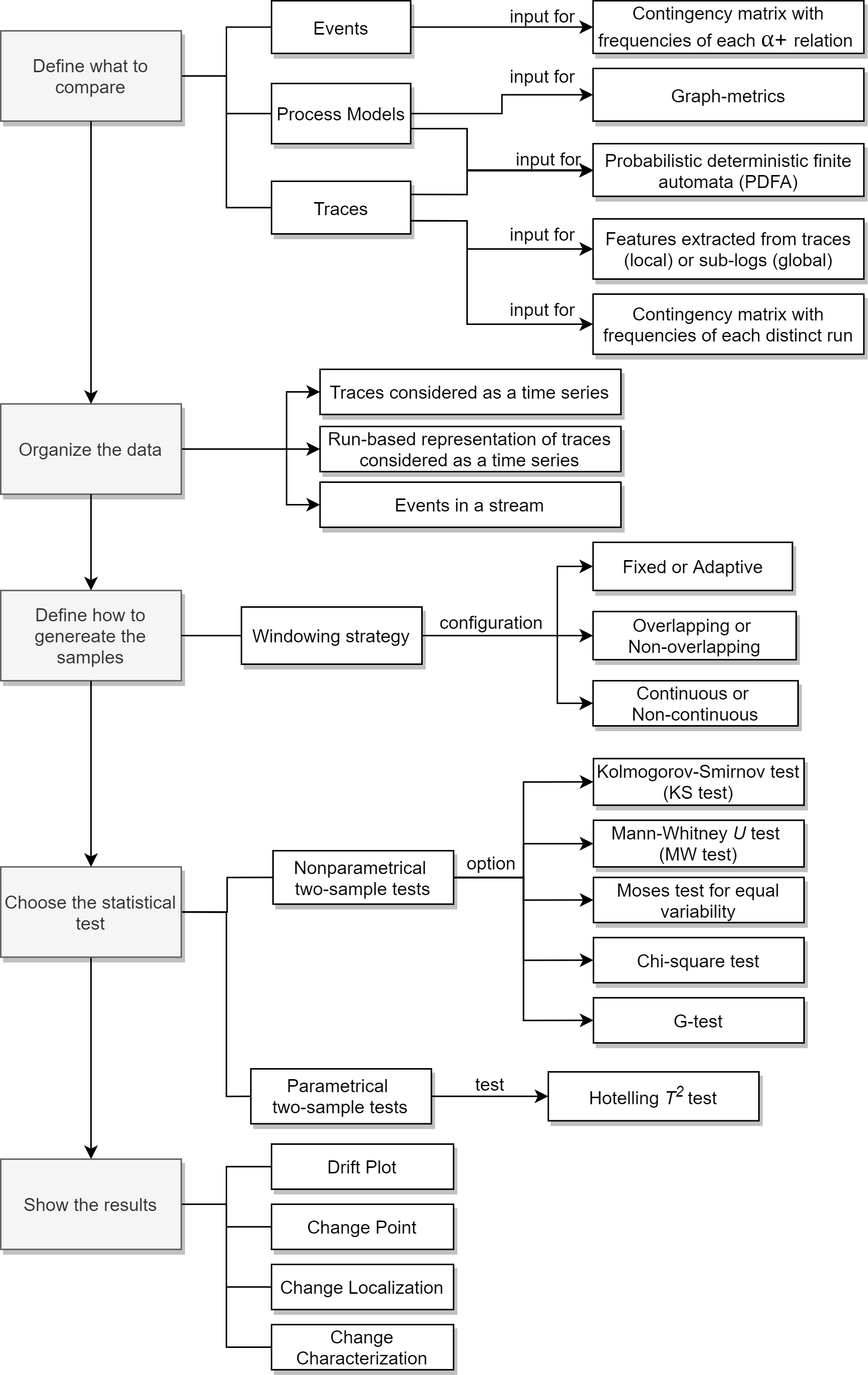}
  \caption{Steps of statistical hypothesis testing approaches to detect concept drift in processes.}
  \Description{The figure identifies the different steps adopted in the papers to apply the standard process of statistical hypothesis testing for drift detection. The main steps are: (i) define what to compare, (ii) organize the data, (iii) define how to generate the samples; (iv) choose the statistical test, (v) show the results, and (vi) evaluate accuracy of the results. }
  \label{fig:statisticalhypothesistesting}
\end{figure}

\paragraph{Define what to compare}

The statistical-based approaches for detecting drifts in processes need to define what information the statistical test should compare. This information can be obtained from event data, discovered process models, or a combination of both. Most of the approaches \cite{Bose2011,Bose2014,Maaradji2015,ManojKumar2015,Martjushev2015,Ostovar2016,Maaradji2017,Ostovar2017,Pauwels2019,Ostovar2020} use a representation of the event data (complete traces or sets of events) to obtain relevant information about them, which is named \emph{features} in some works. In this paper, we use the term \emph{feature} to represent any abstraction calculated from event data or process model. Only one approach uses the actual process model as input for the statistical test. In \cite{Seeliger2017}, the authors obtain graph metrics from discovered process models (discovered applying the Heuristics Miner algorithm \cite{Weijters2006}): (i) number of nodes/edges; (ii) graph density; (iii) in and out-degree of each node; and (iv) occurrence of node/edges. The method summarizes the computed metrics in two vectors (reference and detection) with the size specified by the number of nodes in the graph. For drift detection, the approach uses only the occurrence of edges. The other graph metrics are used to determine the structure of the change.   

In the approaches that use a representation of the event data, we identified several distinct \emph{features} that somehow capture some characteristics from the traces or the events. Authors in \cite{Bose2011,Bose2014} define the concepts of \emph{local} and \emph{global} features. The former is calculated per trace, while the latter is calculated over a log or part of it. Both papers describe the global features relation type count (\emph{RC}) and relation entropy (\emph{RE}); and the local  window count (\emph{WC}) and \emph{J measure} features. One drawback of the proposed method is that the user has to manually pick the feature to be used, implying that one has some \emph{a priori} knowledge about the possible characteristics of the drifts to be detected. Another issue is that the defined features generate a potentially large set of high-dimensional vectors, affecting the scalability of the method for complex real-life logs or even for real-time drift detection \cite{Maaradji2015}. The approach defined in \cite{Martjushev2015} used the \emph{J measure} feature former defined by \cite{Bose2011,Bose2014}, so it shares the same drawbacks aforementioned. Authors in \cite{ManojKumar2015} define one global feature named event class correlation (\emph{ecc}), which indicates if two event classes are linked. A higher \emph{ecc} indicates that the two events commonly happen closely in the traces from the event log. On the other hand, authors in \cite{Maaradji2015,Maaradji2017} proposed a run-based representation of the traces and the notion of run-equivalence, derived from the field of concurrency theory \cite{Van1989}. The set of runs and frequencies are calculated for a given sub-log, indicating this is a global feature. A run can be understood as a representation of a set of traces considering concurrency. 
In \cite{Ostovar2016,Ostovar2017,Ostovar2020}, authors inherit the \(\alpha+\) relations defined by the \(\alpha+\) algorithm \cite{DeMedeiros2004} to generate a contingency matrix containing the frequencies of each identified relation, obtained from sub-logs derived from a set of events. The \(\alpha+\) relations matrix can be considered a global feature because it is calculated over a sub-log containing a set of events, which may contain uncompleted traces. 
The scalability of this feature is validated using one real-life log containing 42 activities and 1,121 traces. Authors in \cite{Pauwels2019} calculates a score for all the cases (after an initial period of training) based on an Extended Dynamic Bayesian Networks (\emph{EDBN}) proposed in the same study. The case score is the mean score calculated overall events within the case and can be classified as a local feature.

\paragraph{Organize the data}

The approaches that use the traces or discovered process models to identify the drifts must organize the input data to generate the samples for applying the statistical test. In the case of comparing complete traces, the methods consider the traces as a time series based on the timestamp of the first event. Next, the features obtained from the traces represent a dataset ordered by the trace arrival time \cite{Bose2011,Bose2014,ManojKumar2015,Martjushev2015,Pauwels2019}. The method applied in \cite{Maaradji2015,Maaradji2017} also organizes the traces as time series by their arrival time. Next, it converts the traces into runs, i.e., a representation that groups together traces with concurrent activities. As the run represents the trace, we consider it a feature obtained from the traces. The process models from \cite{Seeliger2017} are also discovered based on the traces as time series. In contrast, the approach for drift detection applied in \cite{Ostovar2016,Ostovar2017,Ostovar2020} uses an event stream, where the events are emitted according to their timestamp without the need for any particular organization. The features are calculated on a set of events ordered by their arrival time.  

Organizing the event log as a time series may give rise to a tricky situation when dealing with drifts. A case may start with a version of the process, and during its execution, another version can take place (intra-trace drifts). If the traces have a long duration, they can be placed next to each other based on their arrival time, but they can be related to different versions of the process. This situation can confuse the method when identifying the exact point in time where a change begins. 

\paragraph{Define how to generate the samples}
Next, the methods should define how to generate the event samples using a windowing strategy over the dataset of features. The windows can be \emph{fixed} or \emph{adaptive}, \emph{overlapping} or \emph{non-overlapping}, \emph{continuous} or \emph{non-continuous}. The statistical hypothesis test is then applied in two adjacent windows to detect if and when the features significantly change (using a two-sample test), which is the drift detection itself. Usually, the two windows used by the test are named reference and detection windows. The definition of the window size, when set by the user, directly affects the accuracy of the results. The choice of the window size is critical in any drift detection approach because a small window size may lead to false positives, and a large one may lead to false negatives, thus rendering the identification of the drift location challenging \cite{Maaradji2015}. This can be explained by the stability-plasticity dilemma, where the main idea is that learning a concept requires plasticity for the integration of new knowledge, but also stability in order to prevent the forgetting of previous knowledge \cite{Mermillod2013}. Even the adaptive window approach proposed in \cite{Martjushev2015} requires the user to establish a minimum and maximum window size, resulting in false positives or negatives, depending on the values selected. The adaptive window proposed in \cite{Maaradji2015,Maaradji2017} dynamically changes the window size based on the variation observed in the event log, inspired by the ADWIN method proposed in \cite{Bifet2007}. The method needs an initial window size (defined by the user), and despite the authors describe that this parameter can be empirically set, this is not well explored in the paper. When the authors compared the fixed-size window versus the adaptive-size window, they set the initial value for the adaptive approach with the same value of the fixed-window size. Still, there is an open question on how one should define this initial value. In \cite{Ostovar2016}, the authors use the same adaptive window approach without specifying the initial window size. In Table \ref{tab:statistical} we detail the windowing approach for each paper, describing if the window size is a parameter or not. Unfortunately, the authors in \cite{Pauwels2019} do not explicitly describe how the samples are generated.

\paragraph{Choose the statistical test}

The statistical hypothesis testing method's choice depends on the feature (scalar or vector) and the data distribution. Since the event data distribution is not \emph{a priori} known, the authors usually choose non-parametric tests. The only exception is the \emph{Hotelling \(T^2\) test} used in \cite{Bose2011,Bose2014}. The statistical test provides the \emph{p-value}, which is the evidence against the null hypothesis. The null hypothesis indicates that the two samples are equal; i.e., there is no drift between them: the smaller the \emph{p-value}, the more substantial the evidence that the null hypothesis should be rejected. The rejection of the null hypothesis indicates a drift. Because the \emph{p-value} may be under the threshold (indicating a drift) in case of noise, some approaches include a filter to discard abrupt stochastic oscillations in the \emph{p-value}. Authors in \cite{Maaradji2015,Ostovar2016,Maaradji2017} named this filter as the ``oscillation filter'', which allows the method to detect a drift only if a given number of successive statistical tests have a p-value less the typical threshold. The main idea is that a persistent \emph{p-value} under the threshold is much more reliable than a sporadic value happening abruptly.

\paragraph{Show the results}

The most common way to report the results is to plot the \emph{p-value} calculated by the statistical method (often named Drift Plot), with the trace/event index in the x-axis. The Drift Plot may also highlight the CP in the x-axis. The paper \cite{Seeliger2017} only reports the CPs, and the paper \cite{Pauwels2019} plots the case scores instead of the \emph{p-value}, highlighting the drift points.

Table \ref{tab:statistical} reports the details for each of the steps defined in Figure \ref{fig:statisticalhypothesistesting} for the papers using a statistical hypothesis testing approach to detect concept drift in processes. 

Both approaches \cite{Ostovar2017,Ostovar2020} applied the same drift detection mechanism proposed in \cite{Ostovar2016}. The papers are organized by their year of publication and title, using ascending order. 

\subsubsection{Detecting gradual drifts}

Detect gradual drifts are more challenging than sudden drifts because there is no specific point when the cases start emanating from a different version of the process. We identified three statistical-based approaches addressing gradual drifts \cite{Bose2014,Martjushev2015,Maaradji2017}. The method implemented in the Concept Drift plugin \cite{Bose2014} tailored for ProM \cite{VanDerAalst2009} was validated using synthetic logs containing linear gradual changes. The experiments have shown the potential to identify the gradual drifts in the Drift Plot. This approach was extended in \cite{Martjushev2015}, using a noncontinuous sliding window to generate the samples for the statistical hypothesis testing. Therefore, there is a gap between each pair of subsequent windows, allowing for the statistical test to easily identify the drift. The main idea is that the gap will cover when both versions of the process co-exist. The maximum size of the gap has to be manually initialized by the user. This approach assumes the gap period between the samples will cover the time period where both versions of the process coexist. If this assumption is true, a statistical hypothesis test on these two samples should yield a significantly lower \emph{p-value}, thus facilitating the detection of a change. Yet, defining a proper value for the gap parameter is not easy because, depending on how this parameter is set, some drifts may be missed.

Another drawback of this approach is that the detection of sudden and gradual drifts is not integrated, i.e., the user has to choose the type of drift to be detected. When applying the gradual drift detection in event logs containing only sudden drifts, the detected drifts are reported as gradual ones \cite{Maaradji2017}. Thus, it is not possible to use this method to identify the type of the detected drift.   

Another approach that deals with gradual drifts is \cite{Maaradji2017}, implemented in the Apromore Tool \cite{LaRosa2011}. Apromore integrates the detection of sudden and gradual drifts without the need for the user to specify \emph{a priori} the kind of drift he wants to detect. The method relies on the idea that a gradual drift (between \(M_1\) and \(M_2\)) is an interval between two sudden drifts. The first drift indicates the point in time \(M_2\) starts to emanate instances, and the second drift is when there are no more instances from \(M_1\). The approach implements a post-processing stage after the detection of the sudden drifts to identify the gradual drifts. For each consecutive pair of sudden drifts detected, the approach applies another statistical test to check whether the distribution of runs inside the interval is a mixture of runs before the first drift and after the second drift. The authors assume that the distribution in the interval is a linear combination of the distributions before and after the interval and apply the Chi-square goodness-of-fit statistical test to confirm it. This test checks if a set of observations is consistent with a hypothesized distribution. The method uses the histogram of runs as a proxy for the statistical test instead of the full statistical distribution of runs over a given time interval, which is unknown. The validation uses 18 artificial logs that include the gradual drift using a linear probability function with a slope of 0.2 percent.

\begin{landscape}
  \begin{table}[hbt!]
  \caption{Statistical hypothesis testing papers. CP means change point, DP means Drift Plot, WS means window size, and FT means feature.} \label{tab:statistical} 
  \rowcolors{1}{}{lightgray}
  {\begin{tabular}{p{0.9cm}p{1.3cm}p{3.9cm}p{5.5cm}p{4.5cm}p{2.4cm}}
  
  \toprule 
  Paper & Data & Organization & Samples (windowing strategy) & Statistical Test & Results\\\midrule

\cite{Bose2011} & Traces & FTs from traces as time series (local or global) & Non-overlapping, continuous, fixed-size (parameter) & Kolmogorov-Smirnov, Mann-Whitney \emph{U} (univariate data), Hotelling \(T^2\) (multivariate data). & DP (trace) \\

\cite{Bose2014} & Traces & FTs from traces as time series (local or global) & Non-overlapping, continuous, fixed-size (parameter) & Kolmogorov-Smirnov, Mann-Whitney \emph{U} (univariate data), Hotelling \(T^2\) (multivariate data) & Interactive DP (trace)\\

\cite{ManojKumar2015} & Traces & FTs from traces as time series (global) & Non-overlapping, continuous, fixed-size (parameter) & Mann-Whitney \emph{U}, Moses for equal variability & DP (trace)\\

\cite{Martjushev2015} & Traces & FTs from traces as time series (local or global) & Non-overlapping, continuous and non-continuous, adaptive-size (minimum and a maximum WS as parameter) & Kolmogorov-Smirnov & DP (trace) with CP \\

\cite{Maaradji2015} & Traces & Runs as time series (global) & Non-overlapping, continuous, fixed-size (parameter) and adaptive-size (initial WS as parameter) & Chi-square & DP (trace) with CP\\

\cite{Ostovar2016} & Events & \(\alpha+\) relations (events within a window build a sub-log) & Non-overlapping, continuous, adaptive-size (initial WS as parameter) & G-test & DP (trace, event) with CP\\

\cite{Ostovar2017} & Events & \(\alpha+\) relations (events within a window build a sub-log) & Non-overlapping, continuous, adaptive-size (initial WS as parameter) & G-test & DP (trace, event) with CP\\

\cite{Seeliger2017} & Process Models & Graph-metrics as time series & Non-overlapping, continuous, adaptive-size (estimated) & G-test & CPs\\

\cite{Maaradji2017} & Traces & Runs as time series (global) & Non-overlapping, continuous, fixed-size (parameter) and adaptive-size (initial WS as parameter) &  Chi-square, Chi-square goodness-of-fit & DP (trace) with CP\\

\cite{Pauwels2019} & Traces & Case scores based on the EDBN & Not defined & Kolmogorov-Smirnov & Case scores plot with CP\\

\cite{Ostovar2020} & Events & \(\alpha+\) relations (events within a window build a sub-log) & Non-overlapping, continuous, adaptive-size (initial WS as parameter) & G-test & DP (trace, event) with CP\\
  
  \bottomrule
  \end{tabular}}
\end{table}
\end{landscape}

The performance is determined by the \emph{F-score} and \emph{detection delay}. A \emph{TP} is computed if its detected interval includes the central point of the interval of the actual gradual drift. One drawback of this validation scheme is that the distribution of traces during the period when both process models coexisting can be other than linear, which is not validated. However, the method is also applied in a real-life log, the same log used by \cite{Martjushev2015}. The results of the real-life log are validated by the business analyst, confirming the detected drift. In \cite{Maaradji2017}, the authors reported that the business analyst did not recognize the gradual drift reported in \cite{Martjushev2015}, meaning that the detected drift is a FP.

\subsubsection{Change localization}

The methods for dealing with concept drift proposed in \cite{Bose2011,Bose2014} provide an approach for localizing the drifts. The user can interact with the ProM software \cite{VanDerAalst2009}, using the Concept Drift plugin, and select two activities to identify whether there is a drift between them. 

Authors in \cite{Ostovar2017} proposed a fully automated method for characterizing drifts in a process. In the pre-processing step, the approach obtains two sets of data points containing the \(\alpha+\) relations (defined by \(\alpha+\) algorithm \cite{DeMedeiros2004}) from two windows, one before and another after the drift using the drift detection approach from \cite{Ostovar2016}. The size of the data points' set is based on the characterization delay (\(n\)), which is automatically defined to 500, based on performed and reported experiments. In the first stage, the approach performs a statistical test to measure the association between the detected drift and the distribution of the \(\alpha+\) relations before and after it. K-sample permutation test (\emph{KSPT}), as suggested in \cite{Frank1998}, measures the association between each \(\alpha+\) relation (attributes) and the label indicating pre or post-drift (target), to discard irrelevant relations. Next, the method ranks the remaining \(\alpha+\) relations by using the relative frequency change (\emph{RFC}), calculated for each relation. There is also a filter based on the percentage of total relative frequency change (\emph{TRFC}), which considers only a part of the ordered relations, defined as 95\% based on the reported experiments. The second stage of the method matches the considered relations with their RFC with a set of pre-defined change templates and reports the best matches to the user in natural language. The approach was validated using 25 synthetic logs and using a real-life log (BPIC 2011). One limitation is that the method does not characterize simultaneous changes if they have overlapping activities. The characterization delay (\emph{n}) also limits the inter-drifts distance that the method can detect. The authors preferred to provide a fully automated method, not allowing the user to change the parameters, which can also be considered a limitation. Another drawback is that the set of pre-defined templates describes changes in a low level of abstraction, e.g., adding an activity, which results in reporting a lot of low-level changes. The main consequence of the identified limitations is that the drift characterization method hardly works in a real-world environment \cite{Stertz2019}.

In \cite{Ostovar2020}, the authors propose a new method for characterizing the drifts, also based on the drift detection mechanism presented in \cite{Ostovar2016}, to overcome the limitations to handle overlapping and nested changes. First, the approach discovers two process trees from an event stream (pre and post-drift) applying the Inductive Miner – partial traces (IMpt) proposed by the authors. Next, it discovers the sequence of edit operations with minimal cost to convert the pre-drift tree (\(P\)) into the post-drift tree (\(P'\)) by applying a process tree transformation technique. This technique first determines the valid mappings to convert \(P\) into \(P'\), then applies the A* algorithm (described by the authors for the specific application) to find the valid mapping with minimal cost. The authors also implement an option to apply a greedy algorithm to choose the valid mapping, which is faster but provides a non-optimal approach. The implementation is available in the Apromore ProDrift plugin \cite{Ostovar2020}, and it was extensively validated using 365 artificial logs and two real-life logs. 

\subsection{Trace clustering}

Approaches to detect concept drift in processes based on trace clustering analyze different cluster compositions over time. 
The main idea is to consider the time information obtained from traces (or events) in the clustering stage to follow the different compositions of clusters over time. 
New or missing clusters indicate a change in behavior, i.e., concept drift.  
Usually, some traces (or events) are processed as a training step (also named ``grace period'' or ``burnout window'') to determine the default behavior before monitoring different clusters compositions. Sometimes this training step is not explicitly defined; in \cite{Richter2020}, for instance, the authors recommend, as a rule of thumb, neglecting insights from the first \(k\) cases, where \(k\) indicates the size of the hash table used to store the events. 

Authors in \cite{BarbonJunior2018,Tavares2019,Mora2020} proposed the Concept-Drift in Event Stream Framework (CDESF), which monitors the cluster composition over time using time windows to define checkpoints (CPs). When a CP has reached, the information about activities and time intervals from the traces are updated. The method applies a forgetting mechanism to define the traces that should be considered in the update. If the cluster composition changes at the checkpoint, the method triggers a drift. For instance, if the method clusters a case previously identified as anomalous within instances of standard behavior. CDSEF combines frequency of activities and time intervals from traces as features for the clustering strategy, detecting drift in the control-flow and time perspectives combined. Authors in \cite{Accorsi2012} proposed a new clustering strategy applied to the activity distance feature to detect drifts. The strategy checks each trace over time, thus identifying a potential drift when a new cluster is created (cluster cut). The activity distance is calculated between a pair of activities, so a cluster cut between two activities indicates a local drift. The authors proposed a graph showing the combined cluster cuts to provide information about more general drifts to the business analyst. In \cite{Luengo2012}, the authors include the starting time of each process instance as an additional feature used by the clustering approaches. In \cite{Zellner2020}, the authors apply the Local Outlier Factor (LOF) to measure non-conforming traces' distance. Next, traces are aggregated based on a sliding windowing strategy and the calculated LOF. Finally, authors in \cite{Richter2020} propose a monitoring tool based on OPTICS to plot density-based trace clusters in process’ event streams over time. It identifies temporal deviation clusters in a time-dependent graph. We identified the relevant aspects of the approaches for detecting drifts in processes based on trace clustering: defining the features, the clustering strategy, including the time component in the clustering, and the output. Table \ref{tab:trace_clustering} describes each aspect for the identified papers. The papers are organized according to their publication year and title, in ascending order.

\begin{table}[hbt!]
  \caption{Trace clustering considering the time component.}
  \label{tab:trace_clustering}
  
  \resizebox{0.75\textwidth}{!}{
  \rowcolors{1}{}{lightgray}
  {\begin{tabular}{p{1.1cm}p{2.7cm}p{3.2cm}p{3.2cm}p{2.1cm}}
    \toprule
    Paper&Features&Clustering strategy&Time&Output\\
    \midrule
    
    \cite{Luengo2012} & Maximal Repeat Feature Set (\emph{MR}) & Agglomerative Hierarchical Clustering (AHC) \cite{Crc2014} & Timestamp of the 1st event within the trace as a feature. & Clusters\\
    
    \cite{Accorsi2012} & Activity distance & Cluster cuts algorithm (defined by the authors) & After training, each trace is processed by the clustering algorithm & Combined cluster cuts graph \\
    
    \cite{BarbonJunior2018} & Edit and time weighted distances (\emph{EWD}, \emph{TWD}) and time (last event's timestamp) & DBSCAN \cite{Ester1996} & Time interval triggers the clustering, and time composes features for clustering & Flow analysis graph and snapshots of clusters\\
    
    \cite{Tavares2019,Mora2020} & Graph-distance trace and time (\emph{GDtrace}, \emph{GDtime}) & DenStream \cite{Cao2006} & Time interval triggers the clustering, and time is a feature & Drift plot and snapshots of clusters\\
    
    \cite{Richter2020} & Z-scoring of temporal deviation signature (\emph{TDS}) & OPTICS \cite{Ankerst1999} & Events stored in a hash table using the Cuckoo-Hashing & OTOSO plot (time-dependent graph)\\
    
    \cite{Zellner2020} & Local outlier factor (\emph{LOF}) \cite{Breuniq2000}, calculated for non-conforming traces & Aggregate traces with \emph{LOF} scores below (\emph{T}) in one micro-cluster, when the \(n^o\) of traces exceeds \emph{K} & The \emph{LOF} computation is performed in the latest traces by using a fixed-size sliding window & Clusters plotted in a Gantt Chart\\
    
    \bottomrule
  \end{tabular}}
  }
\end{table}

\paragraph{Features}

Trace clustering techniques define a set of features obtained per trace, then group traces with similar features together, calculating the similarity between them using a metric, e.g., Euclidian distance. Thus, the approaches should define how to translate the trace into a feature vector. Authors in \cite{Luengo2012} used the Maximal Repeat Feature Set (MR) \cite{Bose2010}, obtained by counting the maximal repeats in the entire log (concatenating all the traces with a different delimiter) followed by a grouping task of traces with a similar sequence of activities incurring in drift detection on the control-flow perspective. Authors in \cite{Accorsi2012} defined a feature named \emph{activity distance} that indicates the distance between a pair of activities within a trace. This feature is calculated over the whole log as a pre-processing step and allows drift detection on the control-flow perspective. A limitation is that it cannot be used in processes containing loops. The authors in \cite{BarbonJunior2018} defined three features that do represent the trace: edit weighted distance (\emph{EWD}), time-weighted distance (\emph{TWD}), and \emph{Time} (timestamp of the last event of the trace). Both \emph{EWD} and \emph{TWD} are calculated using trace and time histograms. These histograms represent the current behavior in terms of frequency of activities (trace histogram) and time differences (time histogram). Both features indicate the difference between the new trace and histogram information regarding the frequency of activities or time difference. In \cite{Tavares2019,Mora2020}, the authors updated CDESF to use graph-distance trace (\emph{GDtrace}) and graph-distance time (\emph{GDtime}) because histograms do not consider the order relation between activities. CDESF calculates \emph{GDtrace} and \emph{GDtime} comparing the new trace (obtained from the arrived event) and the process graph (\emph{PMG}) normalized. In \cite{Richter2020}, the method uses z-scoring for the temporal deviation signature (\emph{TDS}), calculated based on the mean and variance of all time intervals from each relationship obtained from the cases. The distance between the two traces is calculated using the Euclidean distance. Finally, the authors in \cite{Zellner2020} calculate the (\emph{LOF}) \cite{Breuniq2000}, which describes its ``outlierness'' concerning the surrounding neighborhood. A difference from other approaches is that the LOF is only calculated for non-conforming traces. 

\paragraph{Clustering strategy}

Clustering algorithms group sets of data based on their similarity, maximizing intra-cluster similarity, and minimizing inter-cluster similarity \cite{Crc2014}. They can use different techniques: partitioning, hierarchical, density-based, grid-based, and model-based. Authors in \cite{Luengo2012} apply the Agglomerative Hierarchical Clustering (AHC) \cite{Crc2014} with the minimum variance criterion, using the Euclidian distance between feature vectors. In \cite{Accorsi2012}, the authors define a new algorithm to get the cluster cuts based on the variations of the activity distance (defined as an interval), a feature also defined by the authors. The metric to calculate the similarity is defined by rules based on the four possible interval changes: interval border outrun, smaller interval, no boundary changes, and observation changes. In \cite{BarbonJunior2018}, the authors use DBSCAN \cite{Ester1996}, an algorithm that starts with an arbitrary instance and expands its cluster according to density-based metrics. The algorithm expands regions until all instances are contained in a cluster, or they are considered outliers (an outlier has its density monitored). The main idea is that an increase in the number of instances within the radius of an outlier over time indicates a concept drift. The updated CDESF \cite{Tavares2019,Mora2020} applies DenStream \cite{Cao2006} because it is a density-based clustering method suitable for online scenarios. In \cite{Richter2020}, the proposed monitoring tool applies OPTICS \cite{Ankerst1999} to derive the density-based clusters. Authors \cite{Zellner2020} aggregate traces with similar \emph{LOF} values together without using a previously defined clustering method. They check the traces with a \emph{LOF} score below a threshold \emph{T}, which indicates the affiliation of a trace to a micro-cluster. When the number of traces exceeds a user-given number \emph{K}, these traces are aggregated into a micro-cluster.

\paragraph{Time}

The trace clustering approaches must consider the timestamp of the trace in the cluster definition to detect drifts. In \cite{Luengo2012}, the authors included the timestamp of the first event within the trace as an additional feature on the definition of clusters. In the clustering strategy defined by \cite{Accorsi2012}, each trace is processed as a time series (based on its timestamp), and after each trace, a new cluster may be declared. Authors in \cite{BarbonJunior2018} provided a framework where the amount of anomalous traces can be obtained over time. The time component is firstly considered by retrieving events inside a time horizon (TH) defined by a hyperparameter. By the end of the TH, the framework updates its memory component and triggers the trace and time histograms update. Each case is represented by a triplet [\emph{EWD} – edit-weighted distance, \emph{TWD} – time-weighted distance, \emph{Time} – global time], used in the clustering strategy. \emph{EWD} is calculated using the trace histogram and \emph{TWD} the time histogram. Global time concerns the last event of a given case. The first CDESF version \cite{BarbonJunior2018} also includes time by the feature vector [\emph{EWD}, \emph{TWD}, \emph{Time}]. In the CDESF updated version, the end of the TH triggers the update of the process graph, indicating the current behavior. The tuple [\emph{GDtrace} – trace distance, \emph{GDtime} – trace time distance] represents each case, and it is used in the clustering strategy. Both \emph{GDtrace} and \emph{GDtime} are calculated from graph distance metrics applied between the new processed trace and the current process graph. Then, the feature vector [\emph{GDtrace}, \emph{GDtime}], also considers time for clustering. The Cuckoo-Hashing is applied in the hash table containing cases as a helpful discarding technique for considering the time component in \cite{Richter2020}. The hash table represents a finite set of recent cases, but some older behavior is potentially maintained because the swap operations partially regard the table. The authors in \cite{Zellner2020} include the time component by reading the traces from a stream using a fixed-size sliding window. Gradually, the latest incoming non-conforming traces have the \emph{LOF} calculated.

\paragraph{Output}

The output of this type of approach can be the clusters \cite{Luengo2012}, a plot indicating the CP \cite{Accorsi2012}, or the set of cases representing the concept drift \cite{BarbonJunior2018}. In \cite{Luengo2012}, the clusters are presented, each one indicating a set of traces sharing the same behavior over time. A plot indicating the trace or event where the clustering strategy decides to insert a new cluster is provided in \cite{Accorsi2012}. And the CDESF framework \cite{BarbonJunior2018} plots all the cases considering three monitored dimensions (\emph{EWD}, \emph{TWD}, and \emph{Time}), highlighting the anomalous cases. The clusters that had an increase in the number of samples within the radius of an anomalous one can be identified by the user interaction, indicating a concept drift. In \cite{Tavares2019,Mora2020} CDESF outputs a drift plot and a snapshot of the cluster formation. The OTOSO plot is the output provided in \cite{Richter2020}, which is a time-based graph that allows the detection of different structural changes in an event stream by analyzing the clusters and their connections over time. Authors in \cite{Zellner2020} show the micro-clusters using a Gantt chart. The drifts can be visualized when new micro-clusters appeared in the plot, and the x-axis reports the CP.

CDESF \cite{BarbonJunior2018,Tavares2019,Mora2020} is an approach based on trace clustering that detects concept drifts in the control-flow and time perspectives online. The following aspects support online analysis: the input is an event stream, it handles incomplete traces, it defines a forgetting mechanism, and it is update version uses DenStream clustering algorithm \cite{Cao2006}, which is suitable for online scenarios \cite{Tavares2019}. OTOSO \cite{Richter2020} is also a method for online concept drifts detection in the control-flow perspective. Authors apply a Cuckoo-Hashing forgetting mechanism and handle incomplete traces. They apply OPTICS \cite{Ankerst1999} to derive the clusters and use a time-based graph to show the different structural changes. On the other hand, the approach proposed in \cite{Accorsi2012} is offline. It relies on the user-given window size choice, as a low window size leads to false positives, and larger windows lead to false negatives (undetected drifts). Besides, the method is not designed to deal with loops \cite{Maaradji2015}. The \emph{Dynamic Outlier Aggregation} \cite{Zellner2020} is also classified as offline because it assumes that the streaming events are already gathered to traces in a stream of traces. The validation of CDESF using synthetic datasets does not apply an accuracy metric, e.g., F-score. In \cite{Ceravolo2020}, the authors evaluate CDESF applying MSE and RMSLE. These metrics are not suitable for evaluating concept drift detection accuracy in our understanding because they only indicate if the number of drifts detected is close to the number of actual drifts, while their actual positions are disregarded.

\subsection{Change point detection}

CP detection methods identify the points in which multivariate time series, showing changes in their values \cite{Truong2020}. 
Authors in \cite{Yeshchenko2019,Yeshchenko2019a,Yeshchenko2020,Yeshchenko2021} apply a CP detection method named Pruned Exact Linear Time (PELT) \cite{Killick2012}, which is indicated for datasets with limited size when the number of CPs is not \emph{a priori} known. The Visual Drift Detection (VDD) applies PELT in a pre-defined measure calculated over Declare constraints derived from the log. Firstly, the method mines the log deriving the complete Declare constraints alphabet (\emph{\#cns} is the total of constraints). Next, it splits the log using a fixed-size sliding window and calculates the confidence (or other measure) of each Declare constraint for each window, generating several time series \(T_i=(\textrm{Conf}_{i,1},\ldots,\textrm{Conf}_{i,\#\textrm{win}})\). \(T_i\) contains the confidence value for the constraint i over time (\emph{\#\textrm{win}} is the total of windows). Other metrics can be used, as the support metric, defined in \cite{Yeshchenko2019}. The method derives a multivariate time series \(D=\{T_1,T_2,\ldots,T_{\#\textrm{cns}}\}\) representing the full spectrum of constraints’ confidence.

For drift and CP detection, the method applies the PELT algorithm in \emph{D}, identifying the CPs where a general change in the constraints' behavior occurred. Another option is combining a clustering strategy (hierarchical clustering) for splitting D into groups of constraints with similar confidence trends, then apply PELT. The resulting clusters indicate similar behavior and allow VDD to identify local behavior changes within the clusters. The method reports the detected CPs in one Drift Map (describing the behavior in all clusters) and several Drift Charts (showing the behavior for each cluster). The Drift Chart allows the user to localize the constraints related to change and visually characterize the local drifts \cite{Yeshchenko2019,Yeshchenko2019a} into sudden, gradual, recurring, or incremental (challenges 3 and 4). The user can also inspect if the detected CP has an outlier behavior. In \cite{Yeshchenko2020,Yeshchenko2021}, the tool reported different measures to complement the visual analysis of Drift Charts for identifying the type of the reported drifts. VDD shows the PELT results for sudden drifts, stationarity analysis (Augmented Dickey-Fuller test) for identifying gradual and incremental drifts, and autocorrelation plots for recurring drifts. VDD also allows the visualization of the changes identified in the drifts within the process model. The tool shows the process map (graph) for the complete event log enriched with the different constraints identified for each cluster \cite{Yeshchenko2020,Yeshchenko2021}, providing a mixed visualization.

VDD requires three parameters: window size (to split the event log), window step (value for shifting the sub-log window), and cut threshold (for the clustering strategy). One drawback of this method is that the detected CPs are sensitive to the parameter configuration because the window size determines the granularity of analysis. The method suggests values for the window size and window step in \cite{Yeshchenko2019,Yeshchenko2021}, but this suggestion is based on a good visualization of the plots, not on tuning the accuracy of the tool. In \cite{Yeshchenko2019,Yeshchenko2021}, the authors conclude that the window size does not introduce a significant difference in the results; however, this conclusion is not supported by the performed experiments. 

The evaluation of VDD using synthetic datasets performed in \cite{Yeshchenko2019,Yeshchenko2021} is unclear.
The authors compare VDD with Apromore \cite{Ostovar2016}, but the experiments do not include the complete dataset (datasets are also described using different names), and it is not clear how the authors calculate the F-score. The evaluation using real-world datasets \cite{Yeshchenko2019,Yeshchenko2021} reported that VDD described all types of process drifts comprehensively. Using the BPIC 2011 dataset, the experiment detected all the drifts reported in \cite{Maaradji2017}. All the experiments using real-world datasets in \cite{Yeshchenko2019,Yeshchenko2021} provided an overview of the tool and its user interface, which is complemented in \cite{Yeshchenko2021} by a user study performed with PM experts reporting that the ``visualizations are easy to interact with and useful''. 

\subsection{Visual analysis}

Some authors propose to compare traces (or process models) using a windowing strategy and plotting the differences for visual analysis. Comparing the current time window with the previous one is performed by comparing some specific aspects from the traces without using any statistical test. These works' output is usually a plot with the aspect calculated for a set of traces over time. As a result, the business analyst must analyze the plot to identify the drifts. We have considered studies able to detect drifts in a complete perspective of the process, e.g., control-flow or time. Works comparing a specific attribute, e.g., the time interval between two activities, were not considered.

Authors in \cite{Hompes2017} propose to split the event log based on non-overlapping time windows (by events or minutes) and calculate a similarity matrix for each sub-log. One advantage of this approach is that the matrix can consider any attribute of the event log (from case or event), and an age-decay factor corrects the similarity values. The age factor is applied to reduce the similarity of events apart in time. The differences between adjacent matrices are plotted over time, and the spikes indicate potential drifts. After identifying the drifts, the authors use a clustering strategy to explain them based on previously chosen attributes. This approach is not listed in the clustering approaches because clustering is not used to detect the drift but to analyze the effect of the changing behavior.

The method proposed in \cite{Brockhoff2020} calculates the Earth Mover’s Distance (\emph{EMD}) based on the trace descriptors obtained for sub-logs using a sliding fixed-size window. The approach plots the \emph{EMD} value in a heatmap showing the values for different window sizes. The user can also inspect the \emph{EMD} value in a plot for a specific window size. Based on the \emph{EMD} values, the user can identify potential drifts. A differential of this approach is that \emph{EMD} can be calculated using the control-flow perspective considering frequencies, time perspective (service and sojourn), and a holistic combination of control-flow with time. This drift detection approach allows different perspectives on the process models because of \emph{EMD}'s flexibility regarding the choice of the representation and the distance measure. The approach was experimentally evaluated using one synthetic log, showing that different control-flow types and time-dependent control-flow drifts can successfully be detected. However, further evaluation is needed for applying the approach in real-world scenarios.

In \cite{KurniatiA.P.McInerneyC.ZuckerK.HallG.HoggD.2019}, the authors proposed a manual structured method for concept drift analysis based on the PM2 method \cite{Eck2015}. The method compares processes in three levels: process model, trace, and activity levels. The comparison between process models generates a plot showing replay fitness, precision, and generalization metrics over time. 
Using this plot, users can identify potential drifts (without the exact location). All the metrics are calculated using a process model discovered by the interactive Data-Aware Heuristics Miner (iDHM). In \cite{Kurniati2020}, the authors extended the work applying four miners for performance comparison: Integer Linear Programming (ILP), the interactive Data-Aware Heuristics Miner (iDHM), the Inductive Miner (IM), and the Inductive Miner Infrequent (IMf). The trace and activity analysis can provide information about time and data perspectives but not considering the whole perspective. The authors applied the method to the route to the diagnosis of patients with endometrial cancer over fifteen years. The outputs graphical data visualizations supported discussions about process evolution and changes with domain experts.

\subsection{Change detection}

Authors in \cite{Carmona2012} report the use of a change detection algorithm, ADWIN \cite{Bifet2007}, to identify a drift in event data. The method translates the events into an abstract representation by converting the initial traces from a stream into Parikh vectors \cite{Parikh1966}. From the Parikh vector, it is possible to derive the polyhedron ($P_{\hat{\sigma}}$).
Next, the polyhedra $P_{\hat{\sigma 1}},P_{\hat{\sigma 2}},\ldots,P_{\hat{\sigma k}}$ can be derived from the $k$ traces. Finally, the approach learn the convex-hull of the points represented by the polyhedra $P_{\hat{\sigma 1}},P_{\hat{\sigma 2}},...,P_{\hat{\sigma k}}$, which is a representation of the log (the learned concept). The detection of the drift is then estimated by the ADWIN method, which maintains a window \(W\) of variable size of instances, that is compared at a certain point in time with a previous window. The algorithm automatically grows its window when no change is apparent and shrinks it when data changes. The mass of the polyhedra identifies drifts by updating the \(W\) with 1s (ones) or 0s (zeroes). If the polyhedra (initial concept learned) include the new point observed in the detection stage, its internal window \(W\) is updated with 1, otherwise with 0. When the mass of the polyhedra changes significantly, ADWIN algorithms flags a change with no need to evaluate other parameters. ADWIN also provides rigorous guarantees of its performance, bounding the rates of both false positives and false negatives \cite{Parikh1966}. 

The main drawback of this approach is that it only detects the first drift without identifying the CP. Furthermore, its validation was performed on a small dataset with no considerably set of change patterns. However, this is the first online method to detect concept drifts in processes. This method is also complex and time-consuming since each trace is transformed into multiple prefixes, which are then randomly selected to derive the polyhedron \cite{Liu2018}. Another consideration should be highlighted as, despite being an online method by definition, the validation performed on the paper considers a stream of traces as each trace is converted into several Parikh vectors instead of events. Therefore, to verify the method in an online setting, the Parikh vectors should be created based on an event stream.

In \cite{Hassani2019}, the authors propose StrProMCDD, which applies the ADWIN method \cite{Bifet2007} to detect drift in an event stream. StrProMCDD collects a batch of events in a pruning period (fixed size defined by the user), computes the frequency list for these events, and includes the new frequency list in a temporally ordered list used by ADWIN. A frequency list is a structure containing each pair of activities observed with the respective directly follows relation’s frequency. A directly follows relation indicates that an event follows another event within a trace \cite{VanderAalst2016a}. Because ADWIN builds its observation window based on real numbers, the authors proposed different distance measures derived from the frequency lists: relation frequency distance, dependency and edge distances, activity frequency distance, routing distance, and relative importance distance.

An advantage of StrProMCDD is that it reads events in a stream in a single pass and inherits the time and memory efficiency from the ADWIN method \cite{Bifet2007}. However, the method reads the events in batches, and no validation about the impact of different sizes of these batches is reported. Also, the authors report the F-score metric in \cite{Hassani2019} without clearly defining what is considered as a drift in the method. The plots indicate the size of the adaptive window, and a sharp drop indicates a drift, but it is not clearly defined what threshold triggers a drift. The method also solely addresses the drift detection challenge without reporting the CPs.

Another approach based on a change detection method is proposed in \cite{Impedovo2020}. The authors adapted a pattern-based change detection (PBCD) algorithm, named KARMATree, to detect and characterize drifts on event logs. The events are encoded as a dynamic network to be consistent with the data representation requirements of the PBCD, which represents a sequence of graphs obtained from the traces in the log (treated as time series). Each graph snapshot (\emph{G}) represents a trace (\emph{T}), having an edge for each activity name. One advantage of this representation is that more perspectives of the process model can be included. Yet, the authors only validate the approach using control-flow changes. The authors in \cite{Impedovo2020} validate the accuracy of the KARMATree approach compared with Apromore ProDrift fixed/adaptive using Runs \cite{Maaradji2015}, and ProM adaptive \cite{Martjushev2015} using synthetic logs. The comparison considers F-score and running times with a well-defined experimental protocol (clearly reporting that parameter \emph{minMC} was tuned for the proposed approach). However, neither the synthetic dataset nor the source code is made publicly available.

\subsection{Trend detection}

Authors in \cite{Richter2017,Richter2019} applied a trend detection method called Tesseract adapted from the text mining area for temporal drift detection in event streams. 
Tesseract contains three main parts: (i) monitoring the event stream for collecting the activities’ completion times; (ii) calculating a significance score based on the calculated times and an indicator function; and (iii) visualizing the results. The significance score indicates how far a new observation (time interval between activities) is from the mean value, and it is an adaption from the SigniTrend \cite{Schubert2014} approach (trend detection method from text mining area). However, because of the non-stationary stream environment, the proposed score has exponentially decaying means and variances, and then it is smoothed again (second decay factor) to be a stable indicator. Both decay factors are defined by the user and affect the drift detection. Choosing much smoothing allows the method to rely on significant drift alerts but increases the detection delay and makes it hard to detect short-term anomalies.

Tesseract is implemented as a ProM \cite{VanDerAalst2009} plugin and considers the stream requirements: proposes an adapted Cuckoo hash-table \cite{Fan2014} as a data structure with fast performance for look-up, updating, and deleting; controls memory consumption; deal with a stream of events; minimizes the detection delay. A limitation of Tesseract is that the approach only detects drifts in the time perspective. The reported experiments validate the approach using synthetic and real datasets, indicating that Tesseract is robust to noise and can detect sudden, incremental, and recurring drifts (only by visual inspection). The authors do not evaluate the accuracy of the method using a metric like the F-score. They evaluate the drift in the synthetic dataset by calculating the detection delay, which is the number of events between the start of the drift and the event that triggers the detection.

The Tesseract is not classified as a visual analysis approach because it adapts a trend detection method to automatically triggers the drifts when the significance score is of a minimum number of events shift the value out of the in-control limits, addressing both drift and CP detection. The output of the method is a Control chart with Tesseract values and a Gantt-Chart that plots the Tesseract values exceeding the thresholds, indicating the temporal drifts. The user has to select a pair of activities to visualize both plots, which also provides the change localization of the drift.

\subsection{Other approaches}

Some authors propose different approaches to deal with concept drift in processes that do not fit into any of the previous categories. These approaches are detailed below.

\subsubsection{Tsinghua Process Concept Drift Detection (TPCDD)}

TPCDD \cite{Zheng2017}, is a three-stage approach based on two process similarity algorithm, TAR \cite{Zha2010} and BP \cite{Weidlich2011}, and a clustering strategy. TPCDD handles both drift and CP detection. First, TPCDD creates a relation matrix containing all relations on the traces in the lines and one column for each trace of the event log. This is a limitation of the method to handle complex and large logs. In the second stage, each line of the matrix is inspected to identify intervals containing a stable same frequency level, classified in always, never, and sometimes intervals. The intervals have a minimum size defined by a parameter named minimum relation invariance distance (\emph{MRID}). Each cut between intervals is defined as a candidate CP. In the third stage, TPCDD applies DBSCAN \cite{Ester1996} in the set of candidates CPs to identify the points to be reported as change. DBSCAN parameters: the maximum radius of a neighborhood (\emph{eps}), and the minimum number of points required to form a dense region (\emph{minPts}), are also TPCDD parameters. 

The validation of the approach uses the process models from \cite{Maaradji2015} but with different configurations. The authors generated 32 mixed logs with different models and sizes that were not made publicly available. The accuracy of the method is highly affected by the \emph{MRID} and the DBSCAN main parameters (\emph{eps}, and \emph{minPts}), and the authors did not provide insights on how to define them. Another drawback is that the reported CP indicates the center of the cluster, which can be an invalid trace index. 
The authors described a TP when the CP detected is inside a neighborhood of the precise drift timestamp regarding the F-score metric. 
Yet, by analyzing the plots with the results, the neighborhood is defined by a number of traces defined by the Error Tolerance (\emph{ET}) parameter. Even though the paper describes the validation using the F-score compared to the baseline in \cite{Maaradji2015}, there is a lack of information that does not support the validation process. The authors define two types of relations: direct succession relation (\emph{DSR}) and weak order relation (\emph{WOR}), and TPCDD uses both; however, the experimental protocol did not define which one is applied. The hyperparameters used during the comparison against the baseline (\emph{MRID}, \emph{eps} or \emph{minPts}) were left undefined. We consider this approach relevant as it uses a simplification of the process model (graph) to extract the relations, avoiding the bias of the discovery algorithm. Also, the clustering strategy in the candidate CPs differentiates this approach from the others identified in this survey. However, the variation trend analysis has a lack of statistical foundation. 

\subsubsection{Local Complete-based Drift Detection (LCDD)}

The approach proposed in \cite{Lin2020} applies the local completeness (LC) property of event logs defined in \cite{Yang2012} for drift detection. The idea is that it is possible to assert (with a defined confidence level \emph{K}) that a log satisfies LC in a limited length. Based on this assumption, LCDD defines the minimal length (\emph{ML}), i.e., the minimal number of traces for a log \emph{L} to assure LC with \emph{K}. The detection approach starts by obtaining the \emph{direct succession} (\emph{DS}) relations in a complete window (\emph{CW}), which can be defined based on the calculated \emph{ML} values. Then, it reads traces of the next detection window (\emph{DW}), obtains the \emph{DSs}, and identifies two features: new \emph{DSs} and disappeared \emph{DSs}. A new \emph{DS} or a disappeared one after \emph{CW} in the \emph{DW} indicates a drift. The CW and DW have the same effect as applying sliding windows. LCDD defines the trace with the new \emph{DS} as a CP. If a pre-existent \emph{DS} disappears in the \emph{DW}, LCDD considers the initial trace of \emph{DW} as a candidate CP. LCDD combines the \emph{stable period} (\emph{sp}) to define the exact CP. The value of \emph{sp} indicates the number of traces to be considered a stable period (configured by parameter). After reading \emph{sp} traces, if there are no more disappeared \emph{DSs}, the start of \emph{sp} period is declared as a CP. LCDD also implements an adaptive window based on increasing the complete window size by \emph{minWin} (parameter) in each round.

The authors in \cite{Lin2020} performed several experiments to evaluate the accuracy of LCDD using the \emph{ML} equation to support the definition of the \emph{CW} parameter. The accuracy of the fixed windows obtained using synthetic logs is considerably higher than other approaches: Runs and Alpha options from Apromore ProDrift \cite{Maaradji2015,Ostovar2016}, and TPCDD \cite{Zheng2017}. However, all the compared approaches apply sliding window strategies, but the window size parameter was set to the default value. In LCDD, the \emph{CW} parameter was defined based on the \emph{ML} value. Unfortunately, the papers do not report experiments applying the value calculated for LCDD (200) as the parameter for the compared approaches. There is an interesting statistical foundation in the definition of \emph{ML}, which is a differential of LCDD when comparing it with other fixed sliding window approaches. However, the synthetic datasets used in the validation contain a fixed interval between drifts, which may not represent the real-world behavior. The adaptive approach experiment shows that the accuracy of LCDD is sensitive to the \emph{minWin} parameter. Yet, tuning this parameter can result in higher accuracy when comparing to the default parameters applied to the compared approaches. Authors in \cite{Lin2020} also report a real-world experiment comparing LCDD performance against the same methods applied to the synthetic datasets. Because LCDD proved to be not robust to noise, the authors apply the method defined in \cite{Conforti2017} to obtain similar results to the compared methods. In summary, the accuracy of LCDD is sensitive to the \emph{CW} parameter (but the \emph{ML} equation addressed this drawback), LCDD is not robust to noise, and the accuracy of the proposed adaptive window is sensitive to the parameter \emph{minWin}.

\subsubsection{Framework for ``online'' process concept drift detection}

In \cite{Liu2018}, the authors proposed an approach based on the footprint matrixes containing the \(\alpha\) relations between each pair of activities. First, the method generates a process model from initial traces (a parameter defines the number of traces). Then, the approach checks each new trace by extracting its footprint matrix. Next, the framework compares the matrix of the new trace with the matrix of the current model to identify differences, classified in adding existent or nonexistent activities, removing activities, and altering adjacent relationships between activities. The method applies a metric named process model precision to identify the need to remove process model activities. A difference between the matrixes indicates a drift, and the method returns the activities and the difference to localize and characterize the drift, indicating if the drift is sudden or recurring. There is a filter to consider a trace when it appears more times than a threshold set by the user to avoid noise traces.

Despite being reported as an online method, the definition of an event stream represents a stream of traces. Consequently, the method is not suitable for an online setting. The paper reports that the method is able to identify incremental and gradual drifts, but there are no details about the types of drifts handled by the algorithms or in the validation protocol. Besides, the localization and characterization returned by the algorithms are not validated by the reported experiments. The validation process is not clearly defined; the F-score definition is provided, yet, it is limited for assessment of single-drift scenarios, and the detection error is not explained in the paper. Despite the shortcomings, the approach shows a distinct method to detect drifts based on the footprint matrix and the precision of the process model. 

\subsubsection{Detecting concept drifts based on process histories from event streams}

The approach proposed in \cite{Stertz2018} is based on the discovery of process models providing what the authors named a process history. After reading a new event, the method calculates the fitness of the updated trace (from the case of the new event) for the current model in the process history; a small fitness value (under a pre-defined threshold) triggers the discovery of a new process model, which represents a concept drift. The approach uses an adaptation of the Inductive Miner \cite{Leemans2013} for streams for discovering the process models, based on the Stream-based Process Discovery (S-BAR) architecture defined in \cite{VanZelst2018a}. The proposed algorithm handles memory restriction by delimiting the \(trace_{\textrm{map}}\) structure, which is responsible for storing the \(k\) last traces updated with an event. For conformance checking, the algorithm applies the fitness based on alignments, using the cost for move on log only. One drawback of using the proposed fitness calculation is that if the algorithm discovers a generic model, the fitness will stay with high values, not triggering the insertion of a new process model in the history (no drift will be detected). This is acknowledged by the authors, which propose a periodical model discovery using all traces in the \(trace_{\textrm{map}}\), to detect a stricter model, but this could lead to big mixed process models. Another drawback of this approach is that the validation indicates that there is no initial setup of the process history. When the first event is received from the stream, a new model is mined. As a result, the method identifies incremental concept drifts during the initial events processing, potentially generating false alarms. The method does not report the CPs, just the drift indication and its type (sudden, gradual, incremental, or recurring).

The validation of the method shows that the four types of drift have been detected. Nonetheless, the validation also indicates that the approach misses some concept drifts injected in the synthetic dataset (for the online environment). 
The dataset used contains a small process with simple control-flow changes, and authors do not evaluate the processing time for mining a new model or calculating the fitness after each new event is received. 
Despite the identified drawbacks, authors in \cite{Stertz2018} present an entirely distinct approach to detect drifts based on discovery and conformance for online environments. 
This work is extended in \cite{Stertz2019} to identify concept drifts from the data perspective, validated using a real-world dataset. 

\subsection{Comparing tools for drift detection}
\label{sec:Comparing tools for drift detection}

We identified two papers addressing drift detection without proposing a new technique. They compare existing drift detection tools and report the results. It is essential to highlight that we did not find any paper comparing online PM approaches considering evolving environments.

In the first paper \cite{Omori2019}, the authors compare Apromore ProDrift and the Concept Drift plugin from ProM considering a set of characteristics: detection method, windowing, feature selection, sensitivity, noise handling, type of drift, type of input (log or stream), and user interface. The qualitative analysis of each aspect helps describe both tools. However, there is no accuracy evaluation for the methods based on objective metrics, e.g., F-score. Both methods were applied to a real-world dataset.

Authors in \cite{Ceravolo2020} discuss the properties needed for online PM techniques by defining a set of evaluation goals and evaluate the fulfillment of the defined goals for identified techniques from the state-of-the-art. The paper identified the following goals for online PM: (i) minimize memory consumption, (ii) minimize response latency, (iii) minimize the number of runs, and (iv) optimize accuracy. A set of online process discovery and online conformance checking techniques have been evaluated considering the defined goals. It is essential to highlight that the papers selected in  \cite{Ceravolo2020} do not follow the same criteria defined in our SLR, which is that the online PM approach should report a validation scenario containing drifts. An exciting conclusion of the study proposed in  \cite{Ceravolo2020} is that concept drift detection is a central issue for online PM. However, concept drift techniques are usually not integrated with online PM tasks. This conclusion corroborates with the low number of papers identified on the branch named \emph{online PM dealing with evolving environments}. Because of the importance of concept drift detection for the online PM, the authors compared different concept drift detection tools, limited to open-source software or tools that provide the source code. The paper reports the fulfillment of the goals defined for online PM for each tool and reports a performance evaluation, applying two metrics (from regression methods): mean squared error (\(MSE\)) and root mean squared logarithmic error (\(RMSLE\)). We argue that both metrics are not adequate to measure the accuracy of the drift detection because they consider the presence of the drift but not the CP. So, the conclusions mainly concern the drift detection challenge. The paper publicizes a new dataset including the four drift types identified in the literature (sudden, gradual, incremental, and recurring), articulated according to control-flow and time perspectives \cite{CeravoloStreams2019}.

\section{Online PM dealing with evolving environments}
\label{sec:Online PM dealing with evolving environments}

PM techniques are usually classified into discovery, conformance checking, and enhancement \cite{VanderAalst2016a}. All these tasks can be executed online or offline, and in both situations, concept drift can introduce a challenge. In our SLR, we identified several approaches for detecting concept drifts that can be combined with offline techniques for discovery, conformance checking, or enhancement. When the analyst identifies the drift point, it is possible to split the event log and apply traditional offline techniques on the resulted sub-logs. However, in online analysis, drift detection may not be enough.
For instance, in a process discovery scenario, even if some online drift detection technique reports a drift in an event stream, there still some open issues, i.e., is it possible to mine the model from scratch considering response time limitations? How many events should be used to mine or update the new model? Some papers dealing with concept drift in PM focus on proposing online PM techniques in an evolving environment because of questions like these. We have only found seven papers classified into this branch, and we justified this number of studies (fewer than concept drift detection) because PM research still mainly focuses on offline analysis. Yet, the identified papers only describe process discovery techniques that handle concept drift in online analysis. In the approaches to adapt discovery algorithms after a drift, the challenge may not explicitly detect the drift but enhance the algorithms with adaptive or incremental abilities. We name this challenge as streaming process discovery dealing with evolving environments. We have considered solely the works that validate the streaming discovery approach with a scenario involving concept drifts.

\subsection{Streaming process discovery dealing with evolving environments}

In \cite{Burattin2014}, the authors propose different versions of the Heuristics Miner (HM) \cite{Weijters2006} to deal with event streams, adapting the algorithm using a moving window. Three approaches are designed for evolving environments: HM with Sliding Window (SW), HM with Lossy Counting (LC), and HM with Lossy Counting with Budget (LCB). HW with SW is the simplest approach: first, the process collects events for a given time window, derives an event log, and then applies the classical version of HM. Besides the simplicity, this approach can apply any existing PM algorithm. However, in this approach, only the more recent events are considered, with equal importance. The model update is not constantly triggered, and the algorithm must handle each event at least two times: to store it in the event log and for deriving a model update (not desired for online analysis). The HM with LC applies an adaptation of the standard LC approach \cite{Manku2002}, where the batch version of the directly-follows measure is calculated over an event stream. The algorithm interacts with three data structures: one to count the frequencies of the activities (DA), one to count the frequencies of the direct succession relations (DR), and another one to keep track of distinct cases running at the same time (DC). Because of the concept of ``buckets'' and the cleanup procedure (based on the maximum approximation error allowed) from LC, the algorithm forgets the old behavior. The LC strategy does not limit memory usage, which can be a problem in streaming environments. The LCB \cite{Martino2013} addresses this issue by adapting the approximation error according to the stream and the available budget, i.e., the maximum memory that the algorithm can use. The authors compared the three algorithms using a synthetic event log simulating two concept drifts. The HM with SW reports almost always the worst model-to-model similarity. HM with LC and HM with LCB are basically equivalent, showing the capability to detect the drifts and the corrected model. Experimental results show the effectiveness of control-flow discovery algorithms for streams on a real dataset.

In \cite{Maggi2013}, the authors also propose a framework for the online discovery of declarative process models from streaming event data. Two algorithms for frequency counting are implemented in ProM \cite{VanDerAalst2009}: SW and LC. The authors validate both algorithms using a synthetic log containing sudden drifts. The authors designed three algorithms, each tailored to identify one kind of change template in the Declare constraints: response and not response constraints, precedence and not precedence constraints, and responded existence and not responded existence constraints. The approach is extended in \cite{Burattin2015a}, by including the LCB algorithm for frequency counting. The discovery algorithms are also adapted to apply two different notions of constraint support: event-based and trace-based, and they are now able to discover the entire Declare language. The output of the updated discovery algorithms is also enhanced by providing an updated picture of the process model at runtime (LTL-based business constraints) and information about the most significant concept drifts identified during the process execution. The authors performed different experiments with a larger set of synthetic logs and with a real-life event log (public available). The complete implementation is available in the StreamDeclareDiscovery plugin in the ProM framework \cite{VanDerAalst2009}. Another study \cite{Burattin2015b} complements the approaches mentioned above by focusing on the dynamic visualization of the Declare models over time. The concept drifts can be visually inspected using the trend line plot containing the total of Declare rules discovered over time.

Authors in \cite{VanZelst2018a} proposed a generic architecture, named stream-based abstract representation (\emph{S-BAR}), which allows the adaptation of conventional discovery algorithms for online analysis. 
Any discovery algorithm that maps events into an abstract representation to derive a process model can be adapted by two steps: defining a function for mapping an event log to the abstract representation needed for the algorithm and maintaining a data structure tracking the abstraction's information. 
S-BAR architecture defines some options for the data structure, which should be combined with a forgetting mechanism because of the stream requirements. 
Several instantiations of the S-BAR are implemented in plugins in ProM \cite{VanDerAalst2009}: \(\alpha\)-Miner, Inductive Miner, Heuristics Miner, Transition System Miner (state-based regions), and ILP (language-based regions). The Inductive Miner-based instantiation of S-BAR (based on LC) is validated using an event stream containing one gradual drift with two change patterns. The validation showed that the algorithm is able to adapt the model to the new concept, which is indicated by the replay fitness calculated over time. The experiment also showed that a data structure with more capacity implies the drift being reflected longer.

In \cite{Batyuk2020}, the authors adapted the Fuzzy Miner algorithm \cite{Gunther2007} to the streaming scenario. The implementation of the streaming Fuzzy Miner is integrated into the real-time business process monitoring (RTBPM) software \cite{Batyuk2018}, following the steps: (i) receive and store event into the database; (ii) classify the events; (iii) check if the cases are completed; and (iv) calculate log-based metrics. Specifically, one step of the original Fuzzy Miner is adapted: measuring log-based metrics. The adaptation consisted of organizing calculations to show the changes in a process model in near real-time mode (using as minimal computational resources as possible) to the user. For dealing with drifts, the authors include new metrics: drift unary significance metric, drift binary significance metric, and drift binary correlation metric. The included metrics were calculated between the log-based and derivate metrics (from the original Fuzzy Miner), so the logic for the other metrics remained almost the same. The validation applies the new method to a real-world dataset, comparing the process model generated from the original Fuzzy Miner with the one obtained with the new algorithm. The original Fuzzy Miner derives a process model with an infinite loop, which is not present in the model derived from the streaming Fuzzy Miner. Nonetheless, neither the source code or executable files were made available by the authors \cite{Batyuk2018}.

Another streaming process discovery algorithm is proposed in \cite{Redlich2014}. The authors adapted the Constructs Competition Miner (CCM) algorithm to the streaming environment, considering changes in the behavior. The original CCM applies a divide-and-conquer strategy to mine a block-structured process model from the footprint matrix describing directly-follow relations. In CCM, distinct calculated global relations between activities compete with each other for the most suitable solution. The discovery process also handles noise and not-supported behavior. The proposed Dynamic CCM (DCCM) \cite{Redlich2014}) splits up the original CCM into two parts: (i) footprint update and (ii) the footprint interpretation. For every event, the footprint update module updates the dynamic footprints. DCMM can then compute the dynamic footprints at any point in time be compiled to a business process by the footprint interpretation, scheduled based on a configured number of completed traces. Older behavior is handled by an aging approach applied by the footprint update module. 
It is created an individual trace footprint (TFP) for each trace and adds it multiplied by the trace influence factor \(\texttt{tif} \in R\) to the current dynamic overall footprint (DFP) multiplied by \(1-\texttt{tif}\). 
With this mechanism the influence of a trace completed 60 traces ago became almost irrelevant for \(\texttt{tif} = 0.1\). 
New activities are added to the overall footprint as the trace terminates. Activities that do not appear anymore are removed from the dynamic footprint based on a removal threshold \texttt{tr}. The authors \cite{Redlich2014} validate DCMM in a scenario with a sudden drift in the control-flow perspective. They analyze the change to the business process and report the behavior of the DCCM, the change detection (comparing the trace with actual drift and the detected one), and the change transition period. The reported conclusions indicate that the trace influence factor is a pre-specified value, but it depends on how many traces one must consider representing all the behavior of the process model, which renders the definition of an optimum or even fair value challenging. We did not find any paper comparing the online PM approaches focusing on concept drift scenarios.

\section{Achievements and challenges}
\label{sec:Achievements and challenges}

Drift detection is a hot research topic, and the authors address different challenges: drift detection, CP detection, change localization, change characterization, and revealing the changing process. 
Of the 45 selected papers, 38 of them (approximately 84\%) deal with process drift detection, and the most addressed challenge is CP detection (29 papers). 
Seven papers propose adaptive process discovery techniques for dealing with evolving environments, thus depicting that the PM area is still focused on offline analysis. 
The most addressed problem is CP detection (Figure \ref{fig:countingplots} b) in the control-flow perspective (Figure \ref{fig:countingplots} a). 
Papers addressing more than one perspective are counted for each perspective; the same applies to the challenges plot (Figure \ref{fig:countingplots} b). Regarding the drift perspective, most approaches deal with drifts in the control-flow perspective. The data perspective is explored in \cite{Hompes2017,Pauwels2019,Stertz2019}, and the time perspective is explored in \cite{Richter2017,BarbonJunior2018,Richter2019,Tavares2019,Mora2020,Brockhoff2020}.

\begin{figure}[hbt!]
     \centering
     
     \Description{Plot on the left with the number of papers by perspective: 29 - control-flow; 3 - control-flow and time combined; 3 - time;  2 - data; 1 - control-flow and data combined. Plot on the right with the number of papers by challenge: 29 - CP detection; 14 - change localization; 4 - change characterization; 4 - drift detection (only).}

     \begin{subfigure}[b]{0.47\textwidth}
         \centering
         \includegraphics[width=\textwidth]{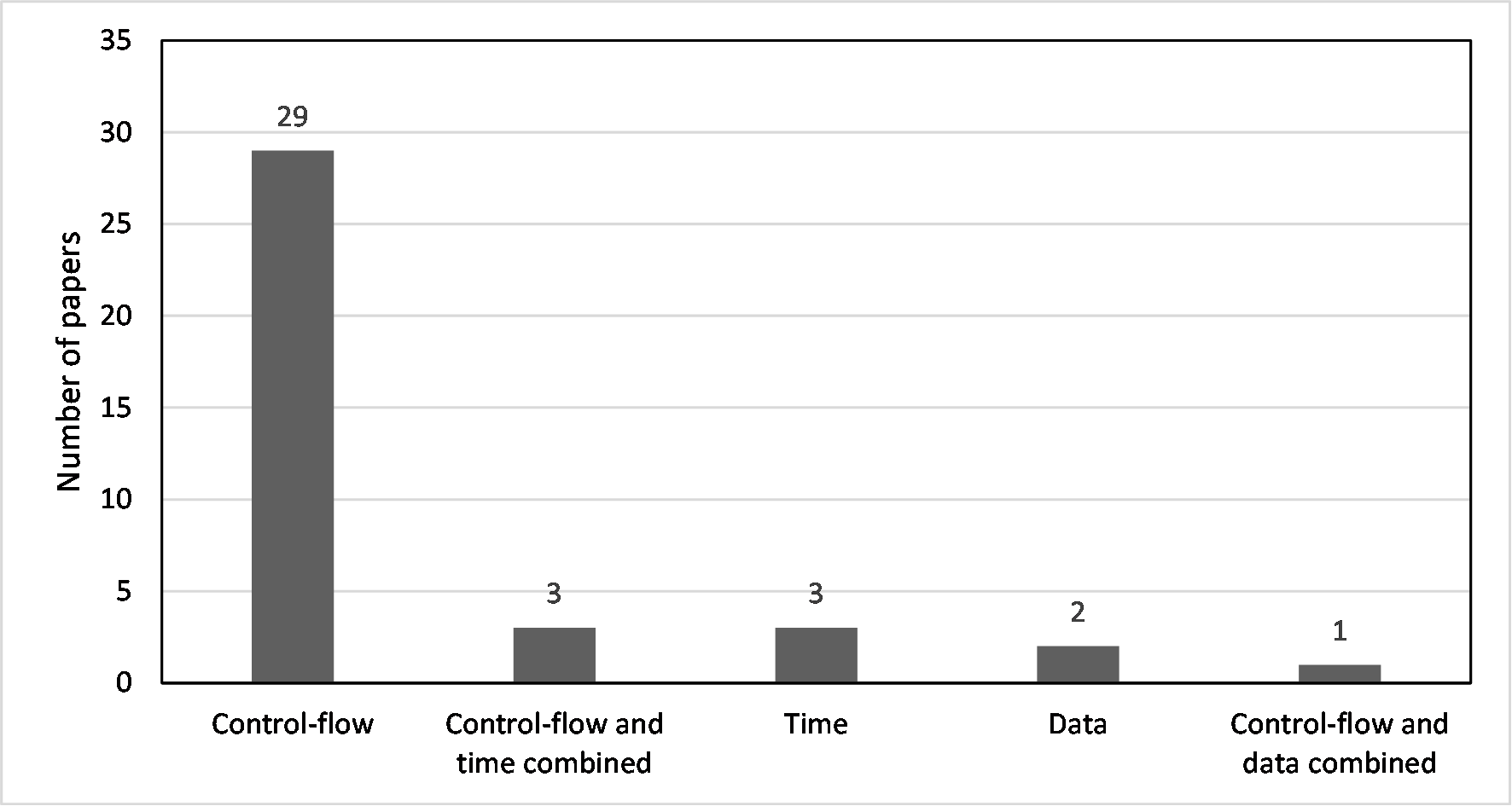}
         \caption{Counted by perspective.}
         
     \end{subfigure}
     \hfill
     \begin{subfigure}[b]{0.47\textwidth}
         \centering
         \includegraphics[width=\textwidth]{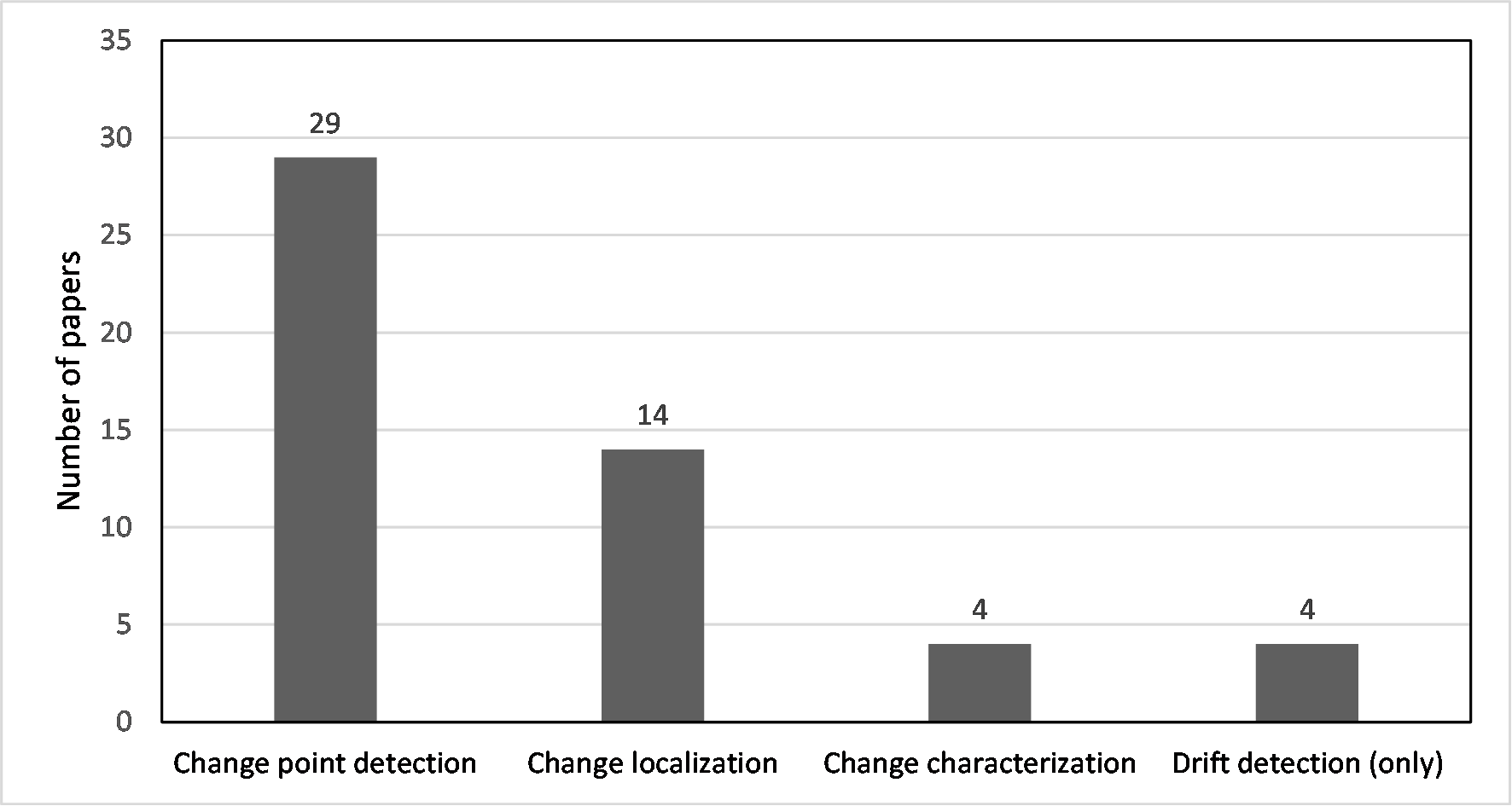}
         \caption{Counted by challenge.}
         
     \end{subfigure}
     \hfill
     \caption{Drift detection papers.}
     \label{fig:countingplots}
\end{figure}

Despite being the most addressed problem, it is difficult to compare the distinct approaches for drift detection in PM. The validation of the detected CP is not always performed using an objective and well-defined metric. Even when the papers report a metric, there are some open questions about how they are computed or regarding the validation protocol. For instance, from the 29 papers that propose a new technique for CP detection, only 14 papers \cite{Hassani2019,Lin2020,Liu2018,Maaradji2017,Maaradji2015,Martjushev2015,Ostovar2016,Ostovar2017,Ostovar2020,Seeliger2017,Yeshchenko2019,Yeshchenko2020,Yeshchenko2021,Zellner2020} use the F-score to validate their results. Yet, the definition of what a TP, TN, FP, or FN stands for is overlooked or different from the definition provided in other papers. We understand that the F-score metric properly describes the accuracy of the detected CPs if the authors define how the true/false positives are counted. We suggest that the TPs can only score CPs detected after the real CP inside a neighborhood pre-defined. The idea of an Error Tolerance (ET), described in \cite{Zheng2017}, similar to the objective evaluation using a lag period proposed in \cite{Lin2020,Martjushev2015}, can be adapted to define this neighborhood. The main difference between the current proposals is that we argue that the lag period should be considered only after the real drift position and not before it, as drift detection is a reactive process rather than a preemptive one.

The two leading software that provide a tool for drift detection, Apromore \cite{LaRosa2011}, and ProM \cite{VanDerAalst2009}, do not report validation metrics in the user interface, e.g., F-score. This hardens the comparison between different methods. We identified only two papers \cite{Ceravolo2020,Omori2019} that compare tools for drift detection in PM. The authors in \cite{Omori2019} do not report a metric for comparing the accuracy of the detected drifts and perform the experiments using a real-life dataset, which is not the best choice when comparing approaches because the real drift points are not known \emph{a priori}. In \cite{Ceravolo2020}, the drift detection accuracy is compared using two metrics applied to regression models (\(MSE\) and \(RMSLE\)). In our understanding, by calculating these metrics, the authors could verify if the approaches detected the correct number of drifts, but the detected drift can be very distant from the actual one. Therefore, we were unable to find any tool available to compare different approaches for drift detection. Providing an easy and common way of comparing different approaches is a challenge in the area. Another challenge is the set of synthetic event logs publicly available. We only identified three datasets (Appendix A) with concept drifts in event logs, and their configuration does not reproduce different drift intervals or a mixed combination of different types of drift expected in real life. The approaches can evaluate their accuracy based on real-world datasets when experts are available for validation to overcome this issue. However, we argue that the tools for generating synthetic datasets should be improved considering changes in processes for future research.

In several approaches for drift detection, the user must set different parameters, and these parameters affect the accuracy of the results. Another challenge is to provide an intuitive and interactive interface, where one can easily change the parameters to tune the results. Some approaches intend to be fully automated, using pre-defined values for the parameters \cite{Maaradji2015,Maaradji2017,Ostovar2016}. However, for different datasets, it should be interesting to provide a way to tune the hyperparameters' values. 

Most papers propose statistical tests over some features extracted from event logs or streams considering the adopted approaches. They apply the statistical tests using fixed or adaptive windows. However, the authors usually propose the adaptive approaches without rigorous guarantees of performance as bounds on the rates of false positives and false negatives. Only two approaches \cite{Carmona2012,Hassani2019} for adaptive windows use the ADWIN method \cite{Bifet2007}, which provides such guarantees. We understand that statistical-based approaches can also enhance adaptive windowing strategies. Also, the application of change detector algorithms (PELT) is reported only in one approach \cite{Yeshchenko2019,Yeshchenko2019a,Yeshchenko2020,Yeshchenko2021}.

The validation of online and offline approaches should follow different experimental protocols. The online setting requires an event stream and has other limitations, e.g., memory consumption and processing time. The experimental protocol of some online approaches only considers the processing time or fails to validate the complete online scenario of analysis. We argue that an event stream is a proper input for online methods because of the restrictions of the online setting, and the experimental protocol should explicitly detail how the method handles: memory, time, and accuracy restrictions. For dealing with concept drifts, the online setting imposes new challenges: the time for drift detection is a critical factor, the number of events needed after the drift occurred is essential, the number of events stored for the analysis must be considered, and the approaches should consider incomplete traces. Besides that, some approaches for online concept drift detection in PM deal with a stream of traces, waiting until the trace is complete and thus, violating the restriction to deal with incomplete traces. We classified these approaches for offline analysis.

\section{Conclusion}
\label{sec:Conclusion}

Despite being a relatively new research topic in the PM community, we found different approaches to deal with concept drift in processes. We categorize the approaches into two main branches: explicitly detect the drift and adapt process mining techniques to deal with event streams in an evolving environment. The primary efforts concern the detection of the concept drift, usually handling the drift and the CP detection in the control-flow perspective. Seven studies have been found on stream process discovery dealing with evolving environments. We believe this reflects that the PM area is still focused on offline analysis, especially when the need is for understanding the real processes. The IC defined for the SLR only considers papers with online process mining approaches that evaluate the method using a scenario containing drifts. We identify few synthetic datasets for this type of validation, and real-world datasets are not suitable because we do not know \emph{a priori} if there is a concept drift. We summarize the following challenges for concept drift in PM:
\begin{itemize}
    \item Enhance the experimental protocol. The use of a detailed metric to evaluate the approaches, either for online and offline analysis, is required. Our suggestion is the use of the F-score, yet, following a precise and clear definition on how to count the true/false positives/negatives, considering an error tolerance, and applying the metric in artificial datasets.
    \item Enhance the online experiment protocol, considering the restrictions of the online environment: detection delay, memory usage, the processing time for each event, and input as an event stream, for instance. 
    \item Include validation in datasets without drifts. The methods are highly sensitive to parametrization, which may lead to false positives. The validation of different thresholds for avoiding the false positives for the proposed parameter can improve the evaluation and robustness of the proposed methods. 
    \item Develop tools for easily comparing the results of different approaches within distinct datasets. Such tools should provide an easy interface to change the parameters and evaluate results. Also, these tools may provide synthetic datasets. 
\end{itemize}

We are in the fourth industrial revolution, Industry 4.0, where one of the five significant features is automation and adaptation. Business process management (BPM) is one of the industrial information integration methods needed, fulfilling the continuous improvement of business processes \cite{Lu2017}. In the context of Industry 4.0, the business process is even more automated and flexible, increasing the need for tools for monitoring and improving the business process in evolving environments. PM provides powerful tools for analyzing, monitoring, and improving the processes; however, the main state-of-the-art techniques do not focus on online and evolving environments. Dealing with concept drift in PM can improve the benefits of this application in the Industry 4.0 context, thus allowing the application of PM techniques in evolving environments.  

\begin{acks}
    The authors would like to thank the support and funding of this research by CAPES Coordenação de  Aperfeiçoamento de Pessoal de Nível Superior - Brasil (CAPES) - Finance Code 001 under Grant Nos.: 88887.509840/2020-00 and 88887.321450/2019-00.
\end{acks}

\appendix
\section{Appendix - Synthetic datasets description}
\label{sec:appendix1}

\begin{enumerate}

  \item	\textbf{Business Process Drift – \cite{Maaradji2015}} - 75 event logs
    \begin{sloppypar}
    \textbf{Link:} \url{https://data.4tu.nl/articles/dataset/Business_Process_Drift/12712436}
    \end{sloppypar}
    
    \begin{sloppypar}
    \textbf{Type of drift:} sudden. \textbf{Perspective:} control-flow (12 simple and 6 complex patterns).
    \end{sloppypar}
  
    \begin{sloppypar}
    \textbf{Size and interval between drifts (traces):} 2500 (250); 5000 (500); 7500 (750); 10000 (1000).
    \end{sloppypar}
  
    \begin{sloppypar}
    \textbf{Number of drifts:} 9 drifts in each log
    \end{sloppypar}
  \hfill 
  
  \item \textbf{Logs Characterization – \cite{Ostovar2020}} - 375 event logs
    \begin{sloppypar}
    \textbf{Link:} \url{https://drive.google.com/file/d/1xYuai8-HBrCZLSAuZMGPv7IJzOjbEsaY/view}
    \end{sloppypar}
    
    \begin{sloppypar}
    \textbf{Type of drift:} sudden. \textbf{Perspective:} control-flow (65 logs with single changes, 30 logs with composite changes, and 30 logs with nested changes)
    \end{sloppypar}
    
    \begin{sloppypar}
    \textbf{Size and interval between drifts (traces):} 3000 (1000)
    \end{sloppypar}
    
    \begin{sloppypar}
    \textbf{Number of drifts:} 2 drift in each log
    \end{sloppypar}
    
    \begin{sloppypar}
    \textbf{Noise:} two variants with 2.5\% and 5\% noise (inserting random events into the traces)
    \end{sloppypar}
    
    \begin{sloppypar}
    \textbf{Complementary information:} Highly variable logs with trace variability around 80\%.
    \end{sloppypar}
  \hfill 
  
  \item \textbf{Synthetic Event Streams - \cite{Ceravolo2020}} - 942 event streams
    \begin{sloppypar}
    \textbf{Link:} \url{https://ieee-dataport.org/open-access/synthetic-event-streams}
    \end{sloppypar}

    \begin{sloppypar}
    \textbf{Type of drift:} sudden, gradual, incremental, recurring. \textbf{Perspective:} control-flow (16 change patterns), time. 
    \textbf{Size (traces):} 100, 500, and 1000
    \end{sloppypar}
  
    \begin{sloppypar}
    \textbf{Number of drifts.} \textbf{Sudden:} 1 drift injected in the middle of the stream. \textbf{Recurring:} 2 drifts – for streams with 100 traces, cases follow the division 33–33–34 (the initial and the last groups come from the baseline, and the inner one is the drifted behavior); for 500 and 1000 traces, the division is 167–167–166 and 330–330–340.
    \textbf{Gradual:} 1 drift (20\% of the stream represents the transition between concepts).
    \textbf{Incremental:} 2 drifts (an intermediate model between the baseline and the drifted model is required; 20\% of the stream contains the intermediate behavior, so the division was 40–20–40, baseline–intermediate model–incremental drift)
    \textbf{Time drifts.}
    Baseline behavior: the meantime is 30 min, standard variation is 3 min.
    Drifted behavior: the mean and standard variations were 5 and 0.5 min. \textbf{Incremental drift:} the transition state (20\% of the stream) was split into 4 parts where standard time distribution decreases 5 min between them, following the incremental change of time.
    \end{sloppypar}
  
    \begin{sloppypar}
    \textbf{Noise:} 4 variants with 5\%, 10\%, 15\%, and 20\% (removing the first or the last half of the trace)
    \end{sloppypar}
  
    \begin{sloppypar}
    \textbf{Complementary information:} The arrival rate of events fixed to 20 min, the time distribution between events of the same case follows a normal distribution. For time drift, the change affects only the time perspective. 
    \end{sloppypar}
\end{enumerate}

\bibliographystyle{ACM-Reference-Format}
\bibliography{references}

\end{document}